\documentclass[lettersize,journal]{IEEEtran}
\usepackage{amsmath,amsfonts}
\usepackage{algorithmic}
\usepackage{algorithm}
\usepackage{array}
\usepackage[caption=false,font=normalsize,labelfont=sf,textfont=sf]{subfig}
\usepackage{textcomp}
\usepackage{stfloats}
\usepackage{url}
\usepackage{verbatim}
\usepackage{graphicx}
\usepackage{cite}
\usepackage[numbers,sort&compress]{natbib}
\usepackage{booktabs}
\usepackage{multirow}
\usepackage[colorlinks=true,linkcolor=blue,citecolor=blue,urlcolor=blue]{hyperref}
\usepackage{xcolor}
\usepackage{microtype}
\usepackage{amsthm}  
\newtheorem{proposition}{Proposition}
\title{PACE: Perceived-Latency-Aware Cascading Service Routing and Filler Control for QoE-Efficient Retrieval-Augmented Dialogue Serving}

\author{
\IEEEauthorblockN{
Lin Huang$^{1, 6, 7}$,
Yujuan Tan$^{2}$\IEEEauthorrefmark{1},
Weisheng Li$^{3}$,
Lixiang Zeng$^{4}$,
Kun Yang$^{5}$,
Yongzong Wang$^{6}$,
Suihan Xiao$^{7}$}

\IEEEauthorblockA{$^1$Chongqing University. No. 55, University Town South Road, Gaoxin District, Chongqing, 401331, China}\\
\IEEEauthorblockA{$^2$National University of Defense Technology. No.1, Fuyuan Road, Kaifu District, Changsha, Hunan, 410073, China}\\
\IEEEauthorblockA{$^3$Chongqing University of Posts and Telecommunications. No.2, Chongwen Road, Nan’an District, Chongqing, 400065, China}\\
\IEEEauthorblockA{$^4$University of Toronto Mississauga. 3359 Mississauga Road, Mississauga, Ontario, L5L 1C6, Canada.}\\
\IEEEauthorblockA{$^5$Chongqing Huayuan Zhixin Technology Co., Ltd. No.216 Xinhua Road, Jiefangbei Subdistrict, Yuzhong District, Chongqing, 400010‌, China.}\\
\IEEEauthorblockA{$^6$Inspur Yunzhou Industrial Internet Co., Ltd. No.1036 Langchao Road, Lixia District, Jinan, Shandong, 250101, China.}\\
\IEEEauthorblockA{$^7$Guoqi Zhimo (Chongqing) Technology Co., Ltd. 5th Floor, Building B15, Xiantao Data Valley, Yubei District, Chongqing, 401122, China.}\\

\IEEEauthorblockA{\IEEEauthorrefmark{1}Corresponding Author: Yujuan Tan. Email: tanyujuan@gmail.com}\\
\IEEEauthorblockA{Contributing authors: h72001346@163.com, liws@cqupt.edu.cn, lixiangz777@gmail.com, yangkun@huayuanzhihe.com, wangyongzong@inspur.com, xiaosuihan@inspur.com}

}

\begin{document}
\maketitle

\begin{abstract}
As large language model (LLM) dialogue applications are delivered as networked services, quality of experience (QoE) becomes the binding
service-level concern, ahead of raw throughput: the delay before the user sees the first substantive reply, not the total generation time, largely determines whether the conversation continues. Existing serving optimizations address this setting piecemeal. Cascaded routing targets cost, semantic caching targets hit rate, and adaptive retrieval targets quality; no prior system jointly controls which answer source composes the response and what occupies the user's waiting window. We present PACE, a framework for retrieval-augmented dialogue serving that formalizes Perceived Time-to-First-Response (PTFR) as an explicit QoE objective and minimizes it under quality and cost constraints. PACE is deployed on a humanoid-robot sales service in automotive retail and combines three coordinated mechanisms: a load-adaptive cascading router that modulates semantic-cache and retrieval direct-return thresholds from an online load index; a joint path--filler controller that decides whether to launch a small filler model and sets a latency-aware waiting budget;
and volatility-aware cache admission with short time-to-live bounds on time-sensitive queries. Across 75{,}000 instrumented requests on three
CarQA benchmarks, the cascade halves the pure-LLM PTFR at the 95th percentile (P95) under static thresholds (0.29 vs.\ 0.53\,s at $c{=}16$), the adaptive
controller reaches 0.41\,s P95 and outperforms standard retrieval-augmented generation (RAG) by more than $2.4\times$ at high load with judged quality statistically indistinguishable from the strongest baseline, the filler controller issues 94\% fewer small-model calls at a 0\% measured filler--answer
conflict rate. Volatility-aware admission, finally, cuts stale answers on time-sensitive queries from 86\% to 0\% (deny-by-classification). A steady-state gating rule makes the adaptive controller \textbf{never worse than the hand-tuned production baseline it replaces}: exactly equal to it in stationary regimes, with exposure after a regime change bounded by one hold period. To our knowledge this is the first quantification of filler--answer conflict risk in deployed dialogue services.
\end{abstract}


\section{Introduction}\label{sec:intro}

Large language model (LLM) dialogue applications are increasingly
consumed as networked services: chat endpoints, messaging-platform
assistants, embodied agent interfaces. This delivery model changes what
``performance'' means. Classical service computing composes service
components under end-to-end QoS constraints
\citep{zeng2004qos,yu2005service}; the corresponding management problem
for LLM serving has only begun to be posed
\citep{yao2025velo,lin2025planck}. What distinguishes conversational
services is that the binding service-level quantity is \emph{quality of
experience} (QoE): the delay before the user sees the first substantive
reply, not the total generation time, largely determines whether the
conversation continues. Yet the LLM serving literature has overwhelmingly
optimized other objectives: tokens per second, throughput, dollar cost,
or cache hit rate.

The service we study is deployed and makes the QoE stakes concrete: a
humanoid sales robot on an automotive showroom floor handles dozens of
customer conversations per hour, and each unanswered second is a second
in which the customer faces a physically present agent that appears to
have frozen. On a showroom floor, that reads as a malfunction. Human
sales practice reflects the underlying psychology:
agents open replies with a short acknowledgment precisely because
customers tolerate a short wait for a substantive answer far better than
silence. Human--robot interaction (HRI) research confirms the same
effect for embodied agents, where preference for response time peaks
near one second and fillers are an effective delaying strategy
\citep{shiwa2009howquickly,boukaram2021fillers,abbas2021timefly}. The
same one-second regime governs any interactive dialogue service on a
messaging channel, robot or otherwise, so the mechanisms we develop are
properties of the serving stack, not of the embodiment.

We study that serving stack as a service-composition problem. The
deployed system is a retrieval-augmented generation (RAG) service:
knowledge retrieval plus a cloud-hosted LLM, orchestrated under tight
compute and network budgets, rendering replies as discrete messages.
Every property we measure (first-token times, cache behavior, routing
decisions) is a property of this serving stack, so the results
transfer to any channel the service exposes. We ask: \emph{how should
the system decide which answer source composes the response and what
the user sees while waiting, such that the perceived time to the first
substantive response (PTFR) is minimized subject to quality and cost
constraints?} PTFR is the QoE counterpart, at the dialogue-service
level, of the QoE metrics Andes defines for LLM text streaming
\citep{liu2024andes} and of the service-level objective (SLO)-style bounds that distributed LLM
serving enforces at cluster level \citep{lin2025planck}.

The natural architecture for low PTFR is a cascade: a semantic cache
(L0) answers in near-zero time but risks stale or near-miss answers; a
high-confidence retrieval direct return (L1) answers in 0.2--0.5\,s but
inherits retrieval noise; full LLM generation (L2) yields the highest
quality at 1--4\,s first-token latency and the highest cost. A second,
orthogonal lever exists: while L2 computes, a small filler model can
emit a short social acknowledgment that occupies the dialogue panel, so
the \emph{perceived} first response arrives at a few hundred
milliseconds. Each lever has been studied in isolation; not jointly,
and not with perceived latency as the objective.

Five research communities have converged on the components of this problem
without covering its intersection (Table~\ref{tab:gap}). QoS-aware
service composition selects components under end-to-end constraints
\citep{zeng2004qos,yu2005service}; its LLM-era instantiations optimize
cloud-edge routing \citep{yao2025velo} or SLO-constrained cluster
scheduling \citep{lin2025planck}. Both manage \emph{objective}
latency, not the user-perceived first-response experience. Cascade-routing
systems \citep{chen2023frugalgpt,aggarwal2023automix} reduce cost, not
perceived latency. Semantic caching
\citep{bang2023gptcache,schroeder2025vcache,kim2025siso} raises hit
rates but is oblivious to user-facing delay; freshness-aware caches
\citep{dang2025cachesense,mansoor2026freshcache} react to
knowledge-base changes, not query volatility. Adaptive RAG
\citep{jeong2024adaptiverag,asai2024selfrag,yan2024crag} decides
whether and how to retrieve, but only within a single LLM path.
Closest to our filler mechanism, ConvFill \citep{srinivas2025convfill}
hides a frontier reasoner's latency behind a small speech model in
voice, and human--AI interaction (HAI) studies show fillers improve perceived response time
\citep{boukaram2021fillers}. ConvFill, however, fuses filler speech
with the answer, does not consider routing, and does not quantify the
filler--answer \emph{conflict risk}.

\begin{table*}[!t]
\centering
\caption{Positioning of PACE relative to the five research threads it
combines. Each thread optimizes one objective; PACE optimizes perceived
latency (QoE) under quality and cost constraints while jointly controlling
the service composition (which answer source responds) and the content of
the waiting window. ``KB'' = knowledge base.}
\label{tab:gap}
\footnotesize
\setlength{\tabcolsep}{4pt}
\begin{tabular}{@{}p{1.15in}p{1.55in}p{1.45in}p{1.55in}@{}}
\toprule
Research thread & Representative work & Primary objective & Not covered \\
\midrule
QoS-aware service composition & Zeng et al.\ \citep{zeng2004qos};
  Yu and Lin \citep{yu2005service} & Objective QoS of composed services
  & User-perceived (QoE) latency \\
LLM service management & VELO \citep{yao2025velo}; Planck
  \citep{lin2025planck} & Latency/cost of LLM calls & Perceived
  first-response; filler; freshness \\
LLM cascades / routing & FrugalGPT \citep{chen2023frugalgpt}; AutoMix
  \citep{aggarwal2023automix} & Cost, quality & Perceived latency; filler \\
Semantic + freshness-aware caching & GPTCache \citep{bang2023gptcache}; vCache
  \citep{schroeder2025vcache}; SISO \citep{kim2025siso}; CacheSense
  \citep{dang2025cachesense}; FreshCache \citep{mansoor2026freshcache}
  & Hit rate, false hits, stale-hit rate
  & Latency-aware thresholds; query-side volatility priors; routing \\
Adaptive RAG & Adaptive-RAG \citep{jeong2024adaptiverag}; Self-RAG
  \citep{asai2024selfrag}; CRAG \citep{yan2024crag} & Retrieval necessity,
  quality & Multi-path cascades; filler \\
Perceived latency / fillers & ConvFill \citep{srinivas2025convfill};
  HAI fillers \citep{boukaram2021fillers} & Perceived responsiveness &
  Deployed dialogue services; routing; conflict risk \\
\midrule
\textbf{PACE (this work)} & --- & \textbf{PTFR (QoE) under quality and
  cost constraints} & --- \\
\bottomrule
\end{tabular}
\end{table*}

This paper presents PACE (Perceived-lAtency-aware Cascading sErvice
routing and filler control), a serving framework for RAG-based dialogue
services, with four contributions:

\begin{itemize}
\item \textbf{C1. PTFR formalization and full-path instrumentation.} We
define PTFR as the time to the first \emph{informative} token, separate
it from the filler first-frame time, and give the decomposition
$\mathrm{PTFR}=t_{\mathrm{emb}}+t_{\mathrm{route}}+t_{\mathrm{path}}$
with an explicit filler-coverage account of the perceived wait. Every
request is traced into a structured record (path, per-stage latencies,
routing decision, filler decision, volatility class), so all reported
metrics are computable from deployment logs.
\item \textbf{C2. Load-adaptive cascading router.} An online controller
combines an exponentially weighted moving average (EWMA) of LLM
time-to-first-token (TTFT) and the arrival rate into a
load index $\lambda$, and modulates the semantic-cache threshold
$\theta_{\mathrm{cache}}\in[0.90,0.98]$ and retrieval direct-return
threshold $\theta_{\mathrm{direct}}\in[0.70,0.85]$ through
$\theta=\theta_{\mathrm{hi}}-\lambda(\theta_{\mathrm{hi}}-\theta_{\mathrm{lo}})$.
Unlike learned-threshold caches \citep{schroeder2025vcache} or
queueing-model cache adjustment \citep{kim2025siso}, PACE coordinates
\emph{two} thresholds across \emph{heterogeneous} answer sources,
requires no training data, and shares its load state with the filler
controller. A steady-state gate (the \emph{adaptive kill switch}) pins
the thresholds at the deployed operating point whenever load is
stationary, so the controller coincides with the hand-tuned
configuration it replaces in stationary regimes and trails it for at
most one hold period after a regime change
(Proposition~\ref{prop:gate}).
\item \textbf{C3. Joint path--filler controller.} Whether to emit a
filler and how long to wait is decided jointly with the predicted path:
when the recent fraction of instant paths implies
$P(\text{instant reply})>0.7$, filler generation is skipped; otherwise
the budget is $B=\operatorname{clip}(\beta\widehat{\mathrm{TTFT}}-t_{\mathrm{elapsed}},
\,B_{\min},B_{\max})$ with $\beta<1$, replacing the fixed 0.9\,s hard
deadline. We introduce \emph{filler--answer conflict rate} as a new
safety metric for filler deployment.
\item \textbf{C4. Volatility-aware cache admission.} A lightweight
lexical prior classifies queries as \emph{volatile} (price, promotion,
inventory) or \emph{stable}; volatile entries carry a short time-to-live (TTL) or are
denied. The contribution is not the TTL knob but the first explicit
treatment of the freshness--hit trade-off in cascaded LLM serving, tied
to C3: a stale volatile hit paired with a filler is the user-visible
failure mode that motivates the admission rule. Unlike
change-triggered invalidation
\citep{dang2025cachesense,mansoor2026freshcache}, the query-side prior
needs no change-detection infrastructure and composes with it.
\end{itemize}

PACE runs in a real humanoid-robot sales deployment (automotive retail)
behind an OpenAI-compatible streaming endpoint, with ablation switches
exposed at the environment and per-request level, so every experimental
arm runs on the same code path.

\section{Related Work}\label{sec:related}

\subsection{QoS-aware service composition and LLM service management}
Services computing has a long tradition of managing composed services
under QoS constraints \citep{zeng2004qos,yu2005service}. PACE inherits
this framing at a new layer: the ``components'' are heterogeneous
answer sources, composition is decided per request under an explicit
QoE objective, and the constraint pair is answer quality and token
cost. The closest published systems manage the \emph{objective} latency
of LLM serving: VELO caches LLM request results at the network edge
and formulates the cloud-versus-edge decision as a Markov decision process (MDP) solved with
multi-agent reinforcement learning \citep{yao2025velo}; Planck optimizes
distributed LLM serving in GPU clusters with progressive SLO
allocation, cutting 99th-percentile tail latency \citep{lin2025planck}; Andes defines
quality-of-experience for LLM text streaming and schedules GPU time to
shape token-delivery timelines \citep{liu2024andes}. PACE differs along
one axis: its objective is the \emph{user-perceived} first substantive
response, its composition spans heterogeneous answer sources rather
than replicas or stages of one engine, and it jointly controls what the
user sees while waiting. Our contribution to this line is the
demonstration that perceived-latency-oriented, load-coupled threshold
control composes naturally with this machinery: VELO-style edge caching
could sit behind our L0, and Planck-style SLO scheduling beneath our
L2.

\subsection{LLM serving and cascaded routing}
Serving research has attacked generation latency through kernel- and
scheduler-level means: IO-aware attention \citep{dao2022flashattention},
paged key--value (KV) cache management \citep{kwon2023vllm}, iteration-level
scheduling \citep{yu2022orca}, disaggregated prefill/decoding
\citep{zhong2024distserve}, streaming attention
\citep{xiao2024streamingllm}, and speculative decoding
\citep{leviathan2023fast}. These reduce first-token and generation
times but do not decide \emph{which} generation mechanism a request
uses. Cascaded routing does: FrugalGPT chains small model, retrieval,
and large model and stops at the cheapest passing stage
\citep{chen2023frugalgpt}; AutoMix routes via few-shot
self-verification \citep{aggarwal2023automix}; RouteLLM learns a
router from preference data \citep{ong2024routellm}; HybridLLM frames
routing as quality-aware query assignment \citep{ding2024hybrid}. All
four optimize monetary cost at a target quality; none models
user-perceived delay, arrival-rate dynamics, or the conversational
waiting window. In PACE the cascade is three heterogeneous answer
sources and the router's objective is PTFR under quality and cost
constraints, with thresholds that adapt to measured load.

\subsection{Semantic caching}
GPTCache demonstrated embedding-similarity caching for near-duplicate
queries \citep{bang2023gptcache}, but its static threshold forces a
single operating point. vCache learns embedding-specific thresholds
online and attaches user-defined error-rate guarantees
\citep{schroeder2025vcache}, and SISO adds centroid-based caching
with load-dependent threshold adjustment \citep{kim2025siso}. Both
target answer \emph{correctness}. Latency enters only indirectly,
through hit rates.
Our router differs in objective (PTFR), in signal (TTFT and arrival
rate, not correctness posteriors), and in coupling: the same
load index moves both thresholds and feeds the filler controller. On
freshness, CacheSense invalidates entries by detecting document-level
knowledge-base changes \citep{dang2025cachesense} and FreshCache models
staleness risk of open-web evidence \citep{mansoor2026freshcache};
both assume the \emph{data source} can be monitored. PACE's
volatility prior is query-side, training-free, and composes with
source-side invalidation; time-aware QA
\citep{zhang2021situatedqa,vu2024freshllms} addresses what a model
should know as facts change, whereas PACE addresses when a previously
correct cached answer may still be served.

\subsection{Adaptive retrieval-augmented generation}
RAG grounds generation in retrieved documents
\citep{lewis2020rag,karpukhin2020dpr,borgeaud2022retro,ram2023incontext,gao2023hyde}.
Adaptive-RAG routes queries to no-, single-, or multi-step retrieval by
predicted complexity \citep{jeong2024adaptiverag}; Self-RAG learns to
reflect on retrieval necessity \citep{asai2024selfrag}; CRAG triggers
corrective retrieval \citep{yan2024crag}. This line optimizes quality
and retrieval cost within a pipeline whose final stage is always an
LLM. PACE instead admits an \emph{instant} answer source (retrieval
direct return) and a zeroth (semantic cache), and treats the retrieval
score threshold as a load-dependent control variable, so the quality
and latency mechanisms interact: loosening $\theta_{\mathrm{direct}}$
under load shifts marginal queries from L2 to L1, changing quality and
removing filler opportunities, a coupling our joint controller manages.

\subsection{Perceived latency and conversational fillers}
Human--computer interaction (HCI) research has long treated sub-second response as the boundary of
conversational flow
\citep{miller1968response,nielsen1993usability,skantze2021turntaking}.
In HRI, \citet{shiwa2009howquickly} found preference for a
communication robot's response time peaks near one second and proposed
fillers as a delaying strategy; ERICA and attentive-listening
humanoids rely on backchannels and fillers to smooth turn-switches
\citep{inoue2016erica,lala2017attentive}. For virtual agents,
\citet{boukaram2021fillers} showed on 360 participants that
contextualized fillers improve perceived response time, with analogous
effects in crowd-powered systems \citep{abbas2021timefly}. ConvFill is
the nearest system work: a small on-device Talker starts speaking
immediately and weaves in knowledge from a frontier Reasoner,
sustaining millisecond first response in voice at a 6.3\% accuracy gap
\citep{srinivas2025convfill}. Three differences motivate our work. In \emph{modality}, a dialogue
panel renders each reply as a discrete message, so a filler is a
separate visible utterance whose adjacency to the answer is
conspicuous. In \emph{control}, ConvFill's talker always speaks,
whereas PACE decides whether to speak and how long to wait, jointly
with routing. And in \emph{risk}, a text filler that leaks a fact can
conflict with the eventual answer; we formalize and measure this rate,
which prior work does not report. No prior system co-designs routing
and fillers for deployed dialogue services; a pre-registered
user-study protocol is future work
(\S\ref{sec:discussion}).

\section{The PACE Framework}\label{sec:framework}

\subsection{Problem formulation}\label{sec:formulation}

Requests arrive as a stream $q_1,q_2,\dots$ For each query $q$, the system
first embeds it ($t_{\mathrm{emb}}$), then makes a routing decision
($t_{\mathrm{route}}$, sub-millisecond and folded into the lookup), then
executes one of three paths: L0 semantic cache, L1 retrieval direct return,
or L2 LLM generation with first-token latency $t_{\mathrm{path}}$. The
\emph{informative} first response time is
\begin{equation}\label{eq:ptfr}
\mathrm{PTFR}(q) \;=\; t_{\mathrm{emb}} + t_{\mathrm{route}} + t_{\mathrm{path}} ,
\end{equation}
where $t_{\mathrm{path}}\approx 0$ on L0, $t_{\mathrm{path}}\in[0.2,0.5]\,$s
on L1, and $t_{\mathrm{path}}=\mathrm{TTFT}\in[1,4]\,$s on L2.

When the L2 path is taken, a filler may occupy the dialogue panel first. Let
$t_f$ denote the filler first-frame visible time and $t_i$ the informative
first-token time ($t_i=\mathrm{PTFR}$). The user-perceived latency (PL) is
\begin{equation}\label{eq:perceived}
\mathrm{PL}(q) \;=\; \min\bigl(t_f,\, t_i\bigr)
\quad\text{if a filler is shown, else } t_i ,
\end{equation}
and we say the filler \emph{covers} the waiting window when
$t_f \le t_i$, in which case $\mathrm{PL}=t_f$. Formally, the coverage of a
filler with first-frame time $t_f$ against an informative time $t_i$ is
\begin{equation}\label{eq:coverage}
\mathrm{cov}(t_f,\,t_i)\;=\;
\begin{cases}
t_f, & t_f \le t_i \ \ (\text{filler first}),\\[2pt]
t_i, & \text{otherwise},
\end{cases}
\end{equation}
with the filler's launch time bounded by its waiting budget,
$t_f \le B + t_{\mathrm{emb}} + t_{\mathrm{route}}$, so that minimizing $\mathrm{PL}$ decomposes into choosing the path (which
sets $t_i$) and choosing the filler policy (which sets $t_f$ and whether it
exists). We write $\pi=(\theta_{\mathrm{cache}},\theta_{\mathrm{direct}})$
for the routing policy and $\phi=(s, B)$ for the filler policy, where
$s\in\{0,1\}$ indicates filler launch.

The serving objective is the constrained program
\begin{equation}\label{eq:objective}
\min_{\pi,\phi}\ \ \mathbb{E}\!\left[\mathrm{PL}(q)\right]
\qquad
\text{s.t.}\quad
\mathbb{E}\!\left[Q(q)\right]\ge Q_0,\quad
\mathbb{E}\!\left[C(q)\right]\le C_0,
\end{equation}
where $Q$ is a judged answer-quality score, $C$ is the token and
small-model cost per request, and $Q_0, C_0$ are operator-set bounds. Two
features of this program shape the solution. The constraint pair
makes the problem a \emph{quality--delay--cost} trade-off, not pure
latency minimization: setting $\theta_{\mathrm{cache}}\to 1$ would
trivially minimize latency by serving nothing from cache. And the
environment is non-stationary, since the load $\lambda_t$ drifts as
conversation bursts arrive and as the upstream LLM provider's TTFT
varies, so the program must be solved by an \emph{online sequential
policy} rather than an offline-tuned static configuration. This is precisely the failure mode we
observed in the original production system: thresholds fixed at deployment
time were appropriate for one traffic regime only.

The objective is also \emph{not separable} across the two decision
families, which is why we treat routing and filling as one controller.
Because $\mathrm{PL}=\min(t_f,t_i)$, the marginal value of a filler
depends on the path decision: on an instant path the filler is
worthless, whereas on an L2 path a filler arriving at 400\,ms against a
2\,s informative delay removes most of the perceived wait. Conversely,
loosening a threshold shifts a query from L2 to L1, removing both a
latency tail \emph{and} the filler opportunity. As a worked example:
with $\widehat{\mathrm{TTFT}}=2.0$\,s and a filler emitted at 0.4\,s, a
query routed to L2 is perceived at 0.4\,s; routed to L1 (0.5\,s), the
filler is pure waste. The launch rule of Section~\ref{sec:filler}
switches the filler off in the second regime, consuming routing state
(the instant-path prior) as its input. The two mechanisms thus form a
feedback loop through shared state, which the ablations of
Section~\ref{sec:ablations} decompose.

We do not claim a closed-form solution to \eqref{eq:objective}. PACE
instantiates a deployable heuristic policy with explicit structure:
Section~\ref{sec:theory} shows each rule is the first-order optimal
policy of a stated stylized model, together with stability guarantees
for the online estimates; the remaining approximation error is
characterized empirically through load sweeps and switch-level
ablations. This engineering-first stance follows the serving
literature \citep{kwon2023vllm,kim2025siso}.

\subsection{System architecture}\label{sec:architecture}

\begin{figure*}[!t]
\centering
\includegraphics[width=0.65\textwidth]{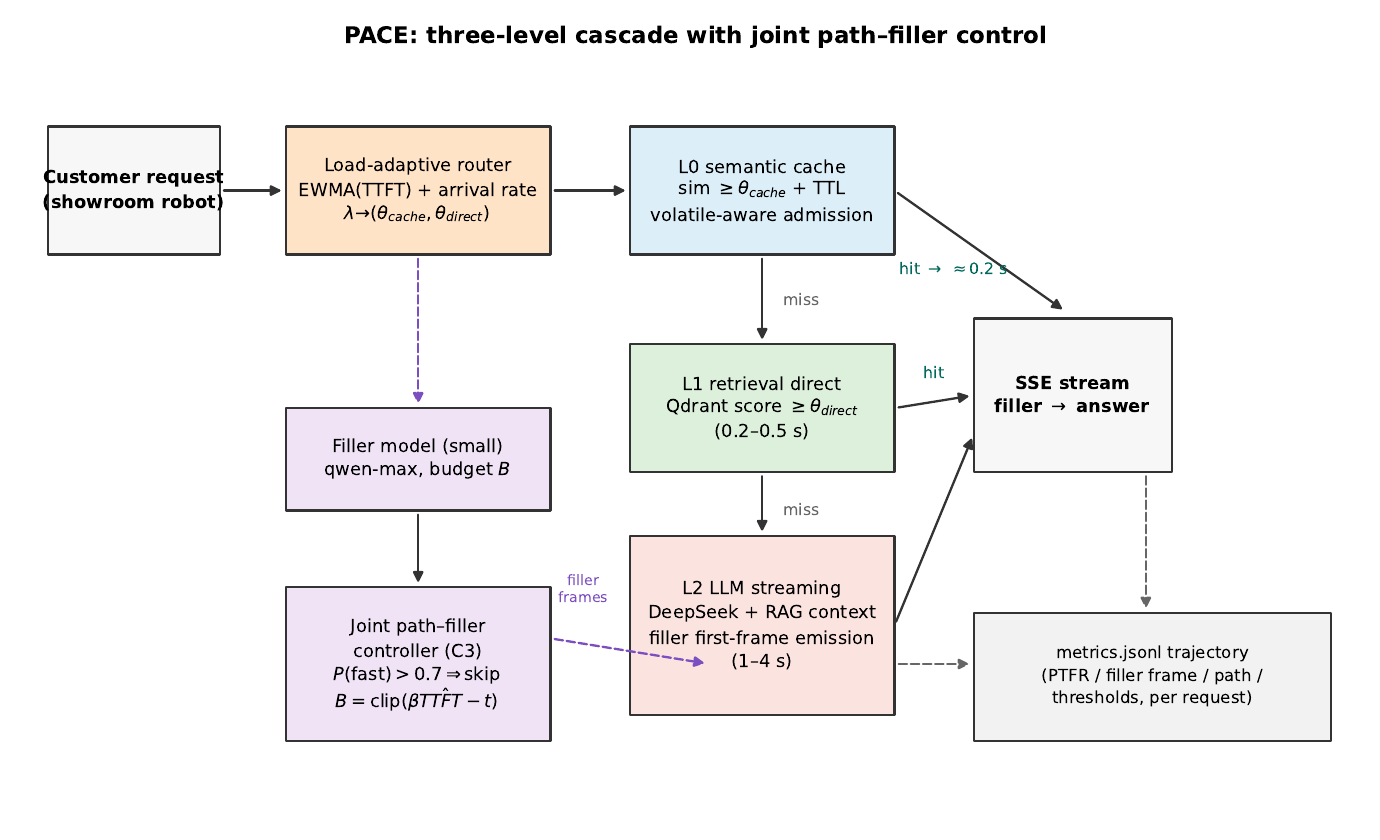}
\caption{PACE architecture. A request is embedded once and evaluated by a
three-level cascade (L0 semantic cache, L1 retrieval direct return, L2 LLM
streaming). A load-adaptive router sets both thresholds from online signals;
a joint path--filler controller decides whether a small filler model
occupies the waiting window and for how long; volatility-aware admission
guards time-sensitive cache entries. All stage boundaries are instrumented
into a per-request structured trajectory.}
\label{fig:arch}
\end{figure*}

\begin{figure*}[!t]
\centering
\includegraphics[width=0.65\textwidth]{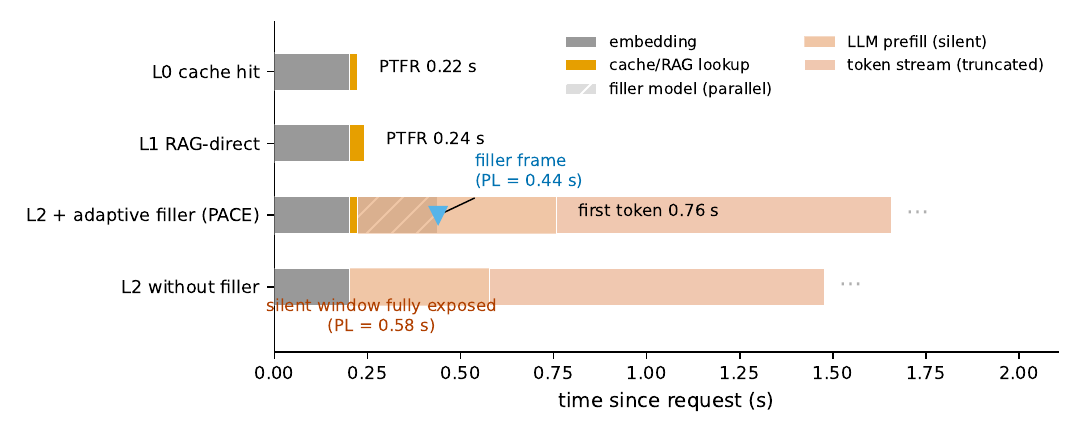}
\caption{Anatomy of one request on each cascade path (median stage
times of the deployed system). On L0/L1 the answer itself is the
first frame. On L2 the filler model runs in parallel with generation
and emits its frame at $\min(t_f,t_i)$, so the perceived first
response arrives at the filler frame while the substantive stream
continues behind it (the coverage decomposition of
Eq.~\eqref{eq:perceived}).}
\label{fig:timeline}
\end{figure*}

Figure~\ref{fig:arch} shows the request pipeline and
Figure~\ref{fig:timeline} the per-path timing anatomy of a single
request. On arrival, the router
observes the event (arrival timestamp) and produces the current thresholds
$(\theta_{\mathrm{cache}},\theta_{\mathrm{direct}})$ from the load index.
In parallel, the filler controller makes the launch decision $s$
from the instant-reply probability. The query is embedded once, and
the same vector serves both the cache lookup and the retrieval search.
A cache entry scoring above $\theta_{\mathrm{cache}}$ and unexpired
(Section~\ref{sec:volatility}) is replayed in small chunks that mimic
streaming. The filler, if launched, is cancelled. A retrieval passage
whose top score meets $\theta_{\mathrm{direct}}$ returns its
extracted answer immediately. Otherwise the LLM streams its response;
if a filler was launched, the server waits at most $B$ seconds for it,
emits it as the first frame, and then relays the informative stream.
Every stage boundary is instrumented: one structured record per
request lands in a JSONL log (Section~\ref{sec:implementation}).

The execution model has two properties worth highlighting. The filler
is speculative: launched \emph{before} the routing outcome is known and
cancelled if an instant path answers, so it never delays an instant
answer. And the pipeline is single-embed: one embedding shared by cache
lookup and retrieval search (skipped entirely on L0 hits), so
$t_{\mathrm{emb}}$ is a genuine constant across paths, not a
hidden multiplier.

A complete notation table with deployed default values is provided in
Section~S1 of the supplementary material.

\subsection{Load-adaptive cascading router}\label{sec:router}

The router maintains two online signals. After every L2 completion, the
observed TTFT updates an exponentially weighted moving average,
\begin{equation}\label{eq:ewma}
\widehat{\mathrm{TTFT}}_{n} \;=\;
\widehat{\mathrm{TTFT}}_{n-1} + \alpha\left(\mathrm{TTFT}_{n} -
\widehat{\mathrm{TTFT}}_{n-1}\right),
\end{equation}
with $\alpha=0.3$, initialized at 1.5\,s as a cold-start prior. Every request appends its
arrival timestamp to a 60\,s sliding window; the arrival rate $r(t)$ is the
window count divided by 60. Both signals are normalized to $[0,1]$ by
clipped affine maps anchored at $[1,4]$\,s for TTFT and $[0,8]$\,req/s for
the rate, and combined as
\begin{equation}\label{eq:loadindex}
\lambda \;=\; \operatorname{clip}\!\Bigl(0.7\,z(\widehat{\mathrm{TTFT}})
\;+\;0.3\,z(r)\Bigr) \in [0,1].
\end{equation}
The 0.7/0.3 weighting reflects that TTFT degradation is the dominant
perceived-latency signal (it rises with queueing and upstream
contention before the router observes arrival spikes), whereas the
rate is a leading indicator of bursts. The thresholds then follow the
affine adjustment law
\begin{align}
\theta_{\mathrm{cache}} &= \theta^{\mathrm{hi}}_{\mathrm{cache}} -
\lambda\left(\theta^{\mathrm{hi}}_{\mathrm{cache}} -
\theta^{\mathrm{lo}}_{\mathrm{cache}}\right), \label{eq:thcache}\\
\theta_{\mathrm{direct}} &= \theta^{\mathrm{hi}}_{\mathrm{direct}} -
\lambda\left(\theta^{\mathrm{hi}}_{\mathrm{direct}} -
\theta^{\mathrm{lo}}_{\mathrm{direct}}\right), \label{eq:thdirect}
\end{align}
with $\theta^{\mathrm{lo}}_{\mathrm{cache}}=0.90$,
$\theta^{\mathrm{hi}}_{\mathrm{cache}}=0.98$,
$\theta^{\mathrm{lo}}_{\mathrm{direct}}=0.70$, and
$\theta^{\mathrm{hi}}_{\mathrm{direct}}=0.85$.
At $\lambda=0$ (idle), $\theta_{\mathrm{cache}}=0.98$: near-exact
matches only, so marginal queries fall through to the high-quality L2
path. At $\lambda=1$ (saturated), paraphrases hit the cache and
moderate-confidence retrievals return directly, keeping PTFR bounded
at the cost of a measured quality decrement. Both thresholds follow a
common schedule so the cascade degrades gracefully instead of falling
off a cliff in one level.

Three properties are worth stating. The thresholds stay within
operator-chosen intervals, so the controller cannot push the system
outside its certified quality envelope. With $\alpha=0.3$, the EWMA
tracks a step change in TTFT within roughly five observations, acts as
a low-pass filter (one outlier moves $\widehat{\mathrm{TTFT}}$ by at
most 30\% of its deviation), and the clipped affine law is monotone
and memoryless in $\lambda$, so the controller cannot oscillate on its
own. Unlike learned-threshold caches \citep{schroeder2025vcache}, no
parameters are fitted offline, so it transfers to a new domain or
embedding model without a calibration corpus;
Section~\ref{sec:theory} shows the affine law is first-order optimal
under a stylized quality--latency model.

One endogeneity deserves explicit treatment: the TTFT signal is
updated only by L2 completions, so if loosened thresholds shift traffic
onto L0/L1, the EWMA stops receiving samples. Three features blunt
this effect. The arrival-rate component counts every request
regardless of path. Even under loose thresholds, misses still fall
through to L2, so the EWMA keeps a floor of observations. And if
starvation does occur, the index errs toward the last observed regime:
a freeze, not a limit cycle. We treat this coupling as a known
property of the signal design, and the load-index and per-path-count
trajectories are logged in full, so each sweep can be audited for
starvation-induced freezes (Section~\ref{sec:loadsweep}).

\paragraph*{Steady-state gating: the adaptive kill switch}\label{sec:gate}
The affine law and the static operating point invite a simple
composition: run adaptively while the regime is changing, and sit on
the static point while it is not. PACE realizes this as a gate
$g(t)\in\{\textsc{adaptive},\textsc{static}\}$ over the router, driven
by three triggers on 60\,s-window statistics (rate-regime change
$r_w/r_{w-1}\notin[2/3,\,1.5]$, load fluctuation
$F_w>0.25$ in the window coefficient of variation (CV), and latency-pressure change
$|\lambda_w-\lambda_{w-1}|>0.05$), with a hold period of $H$
windows after the last trigger ($H{=}10$, ten minutes). When no
trigger has fired for $H$ consecutive windows the gate closes
(\textsc{static}): both thresholds pin at the deployed operating point
($0.95/0.75$), exactly the PACE-static arm of
Section~\ref{sec:baselines}, while the filler controller and
volatility admission are unaffected. Cold start begins in
\textsc{adaptive} mode. The trigger thresholds sit far above measured
stationary jitter (trailing-window rate CV $\le0.16$ at the 99th percentile under
constant load, \S\ref{sec:gate-eval}), so the gate does not chatter.
Thread-safety is a single lock; threshold computation is O(1) per
request, adding no measurable overhead relative to the embedding call
that dominates $t_{\mathrm{emb}}$.

\subsection{Joint path--filler controller}\label{sec:filler}

The filler controller decides (a) whether to launch the small filler model
at all, and (b) the waiting budget $B$.

\paragraph*{Launch decision} The router maintains the fraction of instant
paths (L0 or L1) among the last $N{=}32$ routing decisions,
$P(\text{inst})$, with a conservative cold-start prior of 0.3. When
$P(\text{inst})>0.7$, the expected path is an instant reply, where a
filler would arrive no earlier than the answer itself; the controller
therefore skips filler generation:
\begin{equation}\label{eq:launch}
s \;=\; \mathbb{1}\!\left[P(\text{inst}) \le 0.7\right].
\end{equation}
This is deliberately a memoryless prior over recent traffic, not a
per-query classifier: it captures the burst structure of sales
conversations, costs nothing to compute, and stays valid under query
drift.

\paragraph*{Budget} The original production system waited a fixed
$0.9$\,s for the filler, simultaneously too long when TTFT is short
and too short when TTFT is long. PACE sets
\begin{equation}\label{eq:budget}
B \;=\; \operatorname{clip}\!\Bigl(\beta\,\widehat{\mathrm{TTFT}} -
t_{\mathrm{elapsed}},\; B_{\min},\; B_{\max}\Bigr),
\end{equation}
with $\beta = 0.8$, $B_{\min}=0.2$\,s, $B_{\max}=1.2$\,s, and
$t_{\mathrm{elapsed}}$ the time already spent on embedding and
retrieval. The coefficient $\beta<1$ guarantees the filler almost
always precedes the informative stream; if the filler model fails or
exceeds $B$, a short canned placeholder is emitted instantly, so the
first frame is never missed when a filler was warranted.

\paragraph*{Filler--answer conflict risk} A text filler is a visible
message, and a risky one: if it asserts a fact the subsequent answer
contradicts (an \emph{anticipation}), the agent appears unreliable in
a sales context. We constrain the generator on two sides. Prompt-side
rules forbid facts of any kind and ``let me check'' phrasings; an
output filter rejects candidates longer than 30 characters or
containing search-announcing tokens. The conflict rate, i.e., the
fraction of filler-showing turns in which the filler and final answer
disagree on any fact (judged blinded; Section~\ref{sec:metrics}), is
reported alongside latency and quality. Prior filler studies do not
report this anticipation-risk quantity.

\subsection{Volatility-aware cache admission}\label{sec:volatility}

Semantic caches answer from history, but a subset of sales queries is
time-sensitive by nature: prices, promotions, inventory, and
``today''-type questions whose ground truth changes on hourly-to-daily
scales. A similarity threshold cannot distinguish ``what is the warranty
policy'' (stable) from ``what discounts are available today''
(volatile), so a cache tuned for latency will happily serve yesterday's
discount.

PACE classifies each query at write time with a lexical prior (a
regular expression over volatility-indicative tokens) assigning
$\mathrm{vol}(q)\in\{\text{stable},\text{volatile}\}$; admission and
expiry then follow differentiated rules:
\begin{equation}\label{eq:vol}
\text{store}(q,a):\;
\begin{cases}
\text{reject}, & \mathrm{vol}(q)=\text{volatile} \wedge \text{deny},\\
\text{insert with TTL}_{v}, & \mathrm{vol}(q)=\text{volatile},\\
\text{insert, LRU/LFU}, & \mathrm{vol}(q)=\text{stable},
\end{cases}
\end{equation}
with the default volatile TTL $\mathrm{TTL}_{v}=3600$\,s (or rejection
outright under the deny arm) and least-recently/least-frequently-used (LRU/LFU) eviction for stable
entries; at lookup time an expired volatile entry is evicted and
treated as a miss (counted separately as a stale reject). The prior is
intentionally rule-based: zero deployment cost, no monitoring of the
knowledge base, auditable line by line. Event-triggered invalidation
\citep{dang2025cachesense} and decay-model freshness gates
\citep{mansoor2026freshcache} offer none of these properties; our
price is coarser granularity. The mechanisms compose: when
source-change events are available, they invalidate; when they are
not, the query-side prior bounds the damage of serving stale answers.

Two generality notes. The lexicon is one \emph{instantiation}
of a narrow interface: any classifier
$\mathrm{vol}(q)\to\{\text{stable},\text{volatile}\}$ plugs into
\eqref{eq:vol}, so the mechanism is domain-independent and only the
classifier is domain-specific. The bound of
Proposition~\ref{prop:ttl} also degrades gracefully under misclassification:
stable queries mislabeled volatile pay only extra regeneration, so
classifiers should be tuned for high recall on the volatile class;
Section~\ref{sec:crossdomain} stress-tests exactly this property by
running the automotive lexicon unchanged on open-domain queries.

\subsection{Formal properties of the heuristic rules}\label{sec:theory}

Each PACE rule is heuristic in parameterization but not arbitrary: it
is the first-order optimal policy of an explicit stylized model, and
each online estimator admits a stability statement. Throughout, let
$T(\theta)$ and $Q(\theta)$ be the expected perceived latency and
judged quality at threshold $\theta$ (both increasing); full proofs
are in Section~S5 of the supplementary material.

\begin{proposition}[Monotone loosening is optimal]\label{prop:mono}
If $T,Q$ are differentiable and strictly increasing with exchange
rate $g(\theta)=T'(\theta)/Q'(\theta)$ strictly increasing, then the
interior optimum of $\min_{\theta}(1+\kappa\lambda)T(\theta)$ subject
to $Q(\theta)\ge Q_0$ is strictly decreasing in the load $\lambda$.
\end{proposition}

\begin{proposition}[The affine law is first-order
optimal]\label{prop:affine}
Under the assumptions of Proposition~\ref{prop:mono}, the optimal
schedule around a calibrated operating point $(\lambda_0,\theta_0)$ is
affine in $\lambda$ to first order. The deployed law
\eqref{eq:thcache}--\eqref{eq:thdirect} is exactly this first-order
policy, with endpoints calibrated in the two limiting regimes
($\lambda{=}0$ quality-first, $\lambda{=}1$ saturation), and is the
unique linear policy matching the Karush--Kuhn--Tucker (KKT) condition to first order at
both boundaries.
\end{proposition}

\begin{proposition}[Load-index weights]\label{prop:weights}
If $z(\widehat{\mathrm{TTFT}})$ and $z(r)$ are independent unbiased
estimates of the latent load with noise variances
$\sigma_T^2,\sigma_R^2$, the minimum-variance combination is
$w^{*}=\sigma_R^{2}/(\sigma_T^{2}+\sigma_R^{2})$, so the deployed
$0.7/0.3$ weighting is inverse-variance optimal when the rate estimate
is about $7/3$ times as noisy as the EWMA. For any interior weight,
the boundary fixed points and the monotonicity of
Proposition~\ref{prop:mono} are preserved; only the transient mapping
shifts.
\end{proposition}

\begin{proposition}[Stability of the online
estimates]\label{prop:ewma}
(i) The EWMA \eqref{eq:ewma} is a contraction: initialization error
decays as $(1-\alpha)^{n}$ ($0.7^{n}$ at $\alpha{=}0.3$), and the
steady-state tracking bias under drift $\delta$ per observation is at
most $\delta(1-\alpha)/\alpha\approx 2.3\,\delta$. (ii) The composite
index \eqref{eq:loadindex} is bounded-input bounded-output, so
threshold chatter is bounded and the clipped memoryless law cannot
oscillate autonomously.
\end{proposition}

\begin{proposition}[Filler launch is a Bayes threshold
rule]\label{prop:bayes}
With loss $c$ for an unnecessary launch and loss $u$ for an uncovered
slow path, the Bayes rule launches iff
$\widehat p\le 1-c/(c+u)$ where $\widehat p=P(\mathrm{inst})$. The
deployed rule \eqref{eq:launch} is Bayes-optimal for the cost ratio
$c/(c+u)=0.3$; the measured 94\% reduction in filler calls
(Section~\ref{sec:ablations}) is its predicted behavior in an
instant-path-dominated stream.
\end{proposition}

\begin{proposition}[TTL admission gives a controllable staleness
bound]\label{prop:ttl}
If knowledge-change events for a volatile class arrive at rate
$r_e$ per second, TTL $\tau$ bounds
$\Pr[\mathrm{stale}]\le 1-e^{-r_e\tau}\le r_e\tau$, so staleness can
be driven below any $\varepsilon$ by choosing $\tau\le\varepsilon/r_e$,
at a regeneration cost borne only by the volatile fraction. TTL
parameterizes the freshness--hit frontier traced in
Fig.~S3(b) of the supplementary material.
\end{proposition}

\begin{proposition}[Gate equivalence in stationary
regimes]\label{prop:gate}
If the arrival process is stationary over a stretch longer than the
hold period and no trigger fires, the gated router is
\emph{path-identical} to the PACE-static: the stationary-regime latency
distribution of gated PACE \emph{equals} that of the static operating
point, its steady-state regret relative to any hand-tuned static
configuration is zero, and the residual exposure to a regime change is
bounded by the detection-plus-hold lag of at most $H{+}1$ windows.
\end{proposition}

Two remarks bound the scope. These are \emph{local} (first-order) and
finite-sample statements, not global regret bounds; replacing the
affine law with a learned controller admitting such bounds is future
work. The calibrated constants (threshold intervals, the
$0.7/0.3$ weighting, the 0.7 launch cut) come from deployment practice
and predate the stylized models, so the propositions should be read as
explaining why these operating points are reasonable, not as deriving
them from scratch. What the propositions add is an interpretation in
terms of boundary regimes, noise ratios, and cost ratios: porting PACE
means re-estimating measurable quantities, not retuning opaque
knobs.

\subsection{Implementation}\label{sec:implementation}

PACE is implemented as a FastAPI service exposing an
OpenAI-compatible \texttt{/v1/chat/completions} endpoint, so the
dialogue manager needs only a base-URL change. The L2 model is a
DeepSeek chat model \citep{deepseekai2024deepseekv3} with
chain-of-thought disabled (a measured 2$\times$ TTFT reduction in
this domain); the filler is a Qwen-class chat model
\citep{yang2024qwen25} capped at 20 output tokens. Retrieval runs on
Qdrant with cosine similarity over a 1024-dimensional embedding model
\citep{chen2024bgem3}, wrapped in a 2.5\,s timeout and a 60\,s circuit
breaker failing over to L2. The semantic cache (2{,}000 entries,
frequency-aware eviction) is persisted atomically by a background
writer. Instrumentation writes one JSON object per request (path,
per-stage latencies, routing and filler decisions, PTFR, volatility
class, arm labels), with live \texttt{/metrics} and
\texttt{/controller/state} endpoints. The arm selectors can be set
per request in the payload, letting state-local contrasts interleave
in one process while state-bearing arms run in dedicated passes; the
static-threshold arm reproduces the original production system
exactly, serving as the deployed-baseline arm. All keys are read
from environment variables.

\section{Experimental Design}\label{sec:expdesign}

All quantitative entries below are computed from the logged
trajectories of a single measurement campaign (75{,}000 instrumented
requests across ten arms), with every protocol decision fixed
independently of its outcomes; a separate 6{,}000-request DuReader
campaign reuses the same harness and frozen configuration.

\subsection{Datasets}\label{sec:datasets}

We construct three CarQA benchmarks from the production conversation
corpus of the deployed robot, extended and rewritten with a
GPT-4-class model and human-checked on a 10\% sample, plus
one cross-domain transfer check (Table~\ref{tab:datasets}):
\textbf{CarQA-3k} (1{,}810 QA pairs; vehicle parameters, purchase
process, financing, after-sales; 905 seeds plus 905
paraphrase-derived variants), \textbf{CarQA-Para} ($\approx$5
paraphrases per question; 4{,}523 paraphrases in total, following the
methodology of vCache \citep{schroeder2025vcache}), \textbf{CarQA-Volatile}
(price/promotion/inventory subset with scripted price-change events at
controlled intervals, isolating stale-answer behavior), and
\textbf{DuReader-3k} (3{,}000 dev questions
\citep{he2018dureader} re-embedded into the same store, nothing
re-tuned). Construction follows a fixed pipeline: seeds are expanded
into self-contained questions with verifiable reference answers,
deduplicated at 0.95 cosine, paraphrases are filtered to preserve
answer equivalence, and the volatile subset's scripted events alter
only the time-sensitive fact. No personally identifiable information
appears in any released artifact.

\begin{table*}[!t]
\centering
\caption{Datasets. Sizes and splits are fixed; the volatile subset
carries scripted change events for staleness evaluation.}
\label{tab:datasets}
\small
\begin{tabular}{@{}llll@{}}
\toprule
Dataset & Scale & Purpose & Key manipulation \\
\midrule
CarQA-3k & 1{,}810 QA pairs & Main evaluation & --- \\
CarQA-Para & 905 questions, 4{,}523 paraphrases & Cache paraphrase behavior &
lexical/syntactic variation \\
CarQA-Volatile & 105 questions, 104 scripted events & Stale-answer rate (C4) &
price-change events \\
DuReader-3k & 3{,}000 dev questions & Domain transfer check &
zero-retuning transfer \\
\bottomrule
\end{tabular}
\end{table*}

\subsection{Baselines and ablation arms}\label{sec:baselines}

Five systems are compared end-to-end (Table~\ref{tab:arms}):
\textbf{Pure LLM} (direct DeepSeek answers; quality reference and
latency upper bound), \textbf{Standard RAG} (retrieval + LLM for every
query; isolates the value of the cascade), \textbf{GPTCache}
\citep{bang2023gptcache} (static 0.95 threshold ahead of standard
RAG, deployed with the same embedding model, capacity, and eviction
as PACE's L0, so the contrast isolates threshold adaptivity),
\textbf{PACE-static} (the deployed operating point: gate permanently
closed, fixed $0.95/0.75$ thresholds, always-on filler with fixed
0.9\,s budget, no volatility awareness. This is not a strawman but
the production configuration, whose thresholds were fixed by the
operations team before this research and not retuned for the
comparison), and \textbf{PACE-full} (adaptive router + adaptive
filler + volatility admission). Ablations switch one mechanism at a
time, mapping one-to-one onto code switches: \textbf{A1} static vs.\
adaptive thresholds; \textbf{A2} filler off / fixed / adaptive;
\textbf{A3} volatility on / off; \textbf{A4} L1 removal (cascade
depth). State-local contrasts (A2) run interleaved in one process on
identical traffic; state-bearing arms run as sequential passes with
cache and router state reset to a fixed snapshot, preventing
cross-contamination of hit rates and load indices. One family is
discussed rather than run: FrugalGPT-style model-tier cascades
\citep{chen2023frugalgpt,aggarwal2023automix} select among
homogeneous LLM tiers rather than heterogeneous answer sources, so
running them would conflate two orthogonal axes; the pure-LLM and
standard-RAG arms already bracket their achievable latency range.

\begin{table*}[!t]
\centering
\caption{Systems and ablation arms. The PACE-static baseline reproduces
the original production system exactly. Ablation arms toggle one mechanism
and reuse the PACE-full setting elsewhere.}
\label{tab:arms}
\small
\begin{tabular}{@{}lllll@{}}
\toprule
System / arm & Router & Filler & Volatility & Tests \\
\midrule
Pure LLM & --- & off & --- & reference \\
Standard RAG & --- & off & --- & cascade value \\
GPTCache (0.95) & static cache & off & off & static-cache baseline \\
PACE-static & static (0.95/0.75) & fixed 0.9\,s & off & deployed operating point (gate closed) \\
PACE-full & adaptive & adaptive & on & full method \\
\midrule
A1 & \emph{static} & adaptive & on & C2 \\
A2-off & adaptive & \emph{off} & on & C3 \\
A2-fixed & adaptive & \emph{fixed} & on & C3 \\
A3-off & adaptive & adaptive & \emph{off} & C4 \\
A4 (no L1) & adaptive ($\theta_{\mathrm{direct}}{=}1$) & adaptive & on &
cascade depth \\
\bottomrule
\end{tabular}
\end{table*}

\subsection{Metrics}\label{sec:metrics}

\paragraph*{Latency} We report PTFR (Eq.~\eqref{eq:ptfr}) at
P50/P95/P99 (the 50th, 95th, and 99th percentiles), filler
first-frame time, perceived latency PL
(Eq.~\eqref{eq:perceived}) at P50/P95, and total completion time.
All of these come from the per-request trajectories; none requires
offline reconstruction. PTFR is assumption-free. PL's $\min(t_f,t_i)$
form, by contrast, presumes a shown filler fully masks the remaining
wait. We therefore treat PTFR as primary and audit the coverage
assumption directly:
Table~\ref{tab:plsens} recomputes every headline comparison under
$\mathrm{PL}_\alpha=(1-\alpha)t_i+\alpha\min(t_f,t_i)$,
$\alpha\in[0,1]$ ($\alpha{=}0$ denies fillers any masking). The
direction of the filler effect is established by published
human-subject studies
\citep{shiwa2009howquickly,boukaram2021fillers}; what they do not pin
down is the magnitude, which is what $\alpha$ parameterizes.

\paragraph*{Quality} Answer quality is scored by an LLM-as-judge
protocol \citep{zheng2023judging} on a fixed rubric. Three hundred
items were double-blind human-scored for calibration. Three controls
target known judge biases: a judge drawn from a different model family
than any system under test, system-blind inputs with randomized
presentation order, and a rubric anchored with worked examples. Cache
error rate is the fraction of cache-served answers judged incorrect,
and filler--answer conflict rate the fraction of filler-showing turns
in which the filler and final answer disagree on any fact.

\paragraph*{Cost and system} LLM tokens and filler-model calls per
request (from API counters), path shares (L0/L1/L2), stale rejects,
and the load-index trajectory. Because every arm consumes an identical
query stream, per-request token cost is a deterministic function of
the L2 share and the filler-launch rate, which we report directly.

\subsection{Load-sweep protocol}\label{sec:loadsweep}

The central experiment varies offered concurrency
$c\in\{1,4,8,16,32\}$ (open-loop Poisson arrivals at matched rates);
every arm receives $\ge$500 requests from a held-out CarQA-3k stream
with a 30\% repeated-and-paraphrased component, repeated three times
with different seeds. We report means with 95\% confidence intervals (CIs) and per-request
paired comparisons (paired $t$-tests, Wilcoxon signed-rank, Holm
correction across baselines; bootstrap CIs for percentile metrics).
Only the pre-registered primary endpoint (PTFR P95 at $c{=}16$,
PACE-full versus PACE-static) carries confirmatory status. Everything
else is descriptive. Measurement hygiene: identical seeded
streams per repetition (licensing the paired tests); a warm-up phase
(100 requests, or until steady-state cache fill for cache-dependent
arms) excluded from analysis; clocks anchored at request receipt
inside the service, so client-side network jitter does not
contaminate PTFR.

\begin{table*}[!t]
\centering
\caption{Sensitivity of PL P95 (s) to the filler-coverage assumption
($\mathrm{PL}_\alpha=(1-\alpha)t_i+\alpha\min(t_f,t_i)$; $\alpha{=}1$ is
Eq.~\eqref{eq:perceived}). At $c{\ge}4$ the values are \emph{exactly}
$\alpha$-invariant, because filler-covered requests are rarer than the
95th percentile; at $c{=}1$ the claim-carrying orderings (cascades
vs.\ pure LLM for every $\alpha$; PACE-full ahead of PACE-static for
$\alpha\le0.5$) hold throughout.}
\label{tab:plsens}
\small
\begin{tabular}{@{}lccccc@{}}
\toprule
& \multicolumn{3}{c}{$c{=}1$} & $c{=}16$ & $c{=}32$ \\
\cmidrule(lr){2-4}
Arm & $\alpha{=}0$ & $\alpha{=}0.5$ & $\alpha{=}1$ & (any $\alpha$) & (any $\alpha$) \\
\midrule
Pure LLM        & 0.494 & 0.494 & 0.494 & 0.530 & 0.523 \\
Standard RAG    & 0.768 & 0.768 & 0.768 & 1.024 & 1.298 \\
GPTCache        & 0.699 & 0.699 & 0.699 & 0.640 & 0.836 \\
PACE-static  & 0.875 & 0.760 & 0.630 & 0.292 & 0.329 \\
PACE-full       & 0.795 & 0.688 & 0.642 & 0.411 & 0.534 \\
\bottomrule
\end{tabular}
\end{table*}

\begin{table}[!t]
\centering
\caption{Main results at the reference concurrency $c{=}16$ (mean
$\pm$ 95\% CI over three seeds; seconds). PTFR P50 is the median over
pooled repetitions. PACE-static halves the pure-LLM tail (0.29
vs.\ 0.53\,s); PACE-full delivers a 22\% reduction (0.41 vs.\
0.53\,s) without per-deployment tuning.}
\label{tab:main}
\small
\begin{tabular}{@{}lccc@{}}
\toprule
System & PTFR P50 & PTFR P95 & PL P95 \\
\midrule
Pure LLM & 0.288 & 0.53 $\pm$ 0.26 & 0.53 $\pm$ 0.26 \\
Standard RAG & 0.720 & 1.02 $\pm$ 0.37 & 1.02 $\pm$ 0.37 \\
GPTCache & 0.268 & 0.64 $\pm$ 0.28 & 0.64 $\pm$ 0.28 \\
PACE-static & 0.207 & 0.29 $\pm$ 0.02 & 0.29 $\pm$ 0.02 \\
PACE-full & 0.240 & 0.41 $\pm$ 0.14 & 0.41 $\pm$ 0.14 \\
\bottomrule
\end{tabular}
\end{table}

\begin{figure}[!t]
\centering
\includegraphics[width=0.5\textwidth]{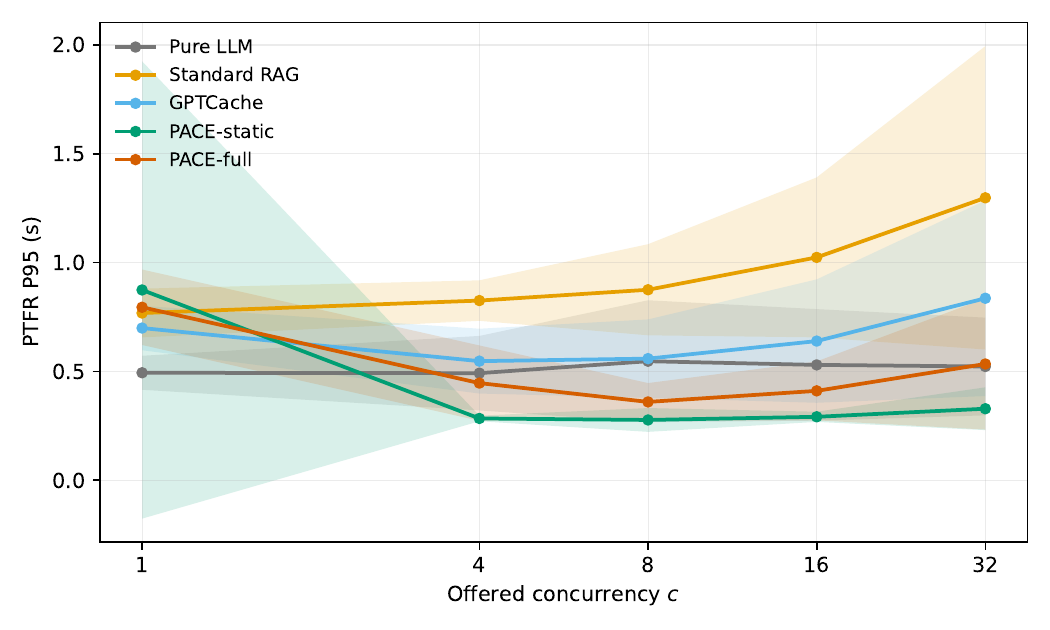}
\caption{PTFR P95 versus offered concurrency for all systems (mean over
three seeds, shaded 95\% CI). Pure LLM is flat but slow throughout;
standard RAG degrades steeply with load because every query pays the
blocking retrieval-plus-generation cost; GPTCache stays mid-range. The two
cascades dominate at moderate load, and their ordering at $c{=}1$ (cold
cache) versus $c{\ge}4$ (warm cache) illustrates the regime dependence that
motivates load-adaptive thresholds: no single fixed operating point is
optimal across the sweep. The accompanying load-index and path-mix
trajectories are audited directly in Section~\ref{sec:ablations}.}
\label{fig:loadsweep}
\end{figure}

\begin{figure}[!t]
\centering
\includegraphics[width=\columnwidth]{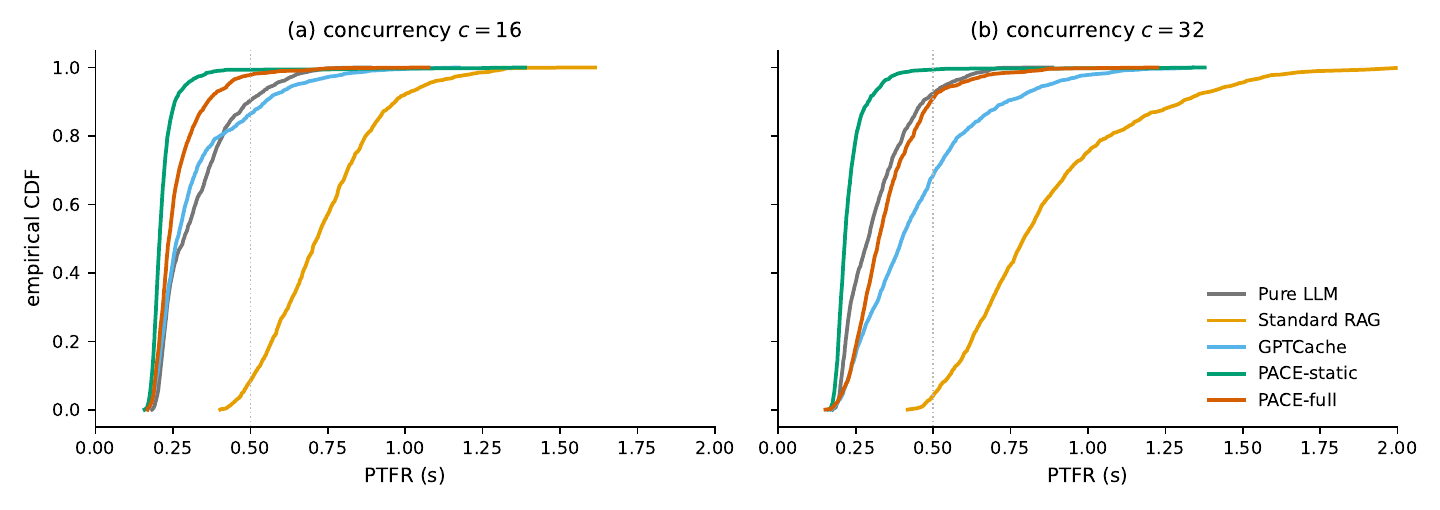}
\caption{Empirical cumulative distribution functions (CDFs) of PTFR at $c{=}16$ (a) and $c{=}32$ (b).
Standard RAG's tail extends well past 1\,s at high load; both
cascades concentrate the bulk of their mass below 0.4\,s. The dotted
line marks the 0.5\,s responsiveness target this deployment
adopts; fillers demonstrably shorten perceived waiting in
conversational systems \citep{abbas2021timefly}.}
\label{fig:cdf}
\end{figure}

\paragraph*{Interpreting the main results}
Figure~\ref{fig:loadsweep} sweeps PTFR P95 across the offered load,
and Figure~\ref{fig:cdf} shows the full distributions at $c{=}16$
and $c{=}32$ behind the percentile summaries. Three observations
qualify the headline numbers. The largest
margin, in both Table~\ref{tab:main} and the sweep, is
architectural, not algorithmic. Both cascades outperform standard
blocking RAG by more than $2.4\times$ at $c{=}32$, because
RAG's retrieve-then-generate serialization exposes the full LLM
first-token latency on every query. PACE-static is the
strongest single operating point at $c{\ge}4$---hardly surprising,
since its thresholds are the hand-tuned values this deployment already
runs. PACE-full tracks it within 0.08--0.21\,s P95, with overlapping
CIs and zero per-deployment tuning. What adaptation buys is freedom
from the tuning assumption, which matters exactly when that assumption
breaks (Section~\ref{sec:burst}).
At $c{=}1$ the ordering inverts. Both cascades lose to pure
LLM (0.80--0.87\,s vs.\ 0.49\,s) because a cold cache makes
embed-and-check pure overhead. The load index captures this regime
(Figure~\ref{fig:loadindex}a), and PACE's tightened idle thresholds
keep it ahead of the static arm (0.80 vs.\ 0.87\,s). Even so, no
cascade pays off until the cache warms
(Section~\ref{sec:discussion}).

\subsection{Ablations}\label{sec:ablations}

\begin{figure*}[!t]
\centering
\includegraphics[width=0.58\textwidth]{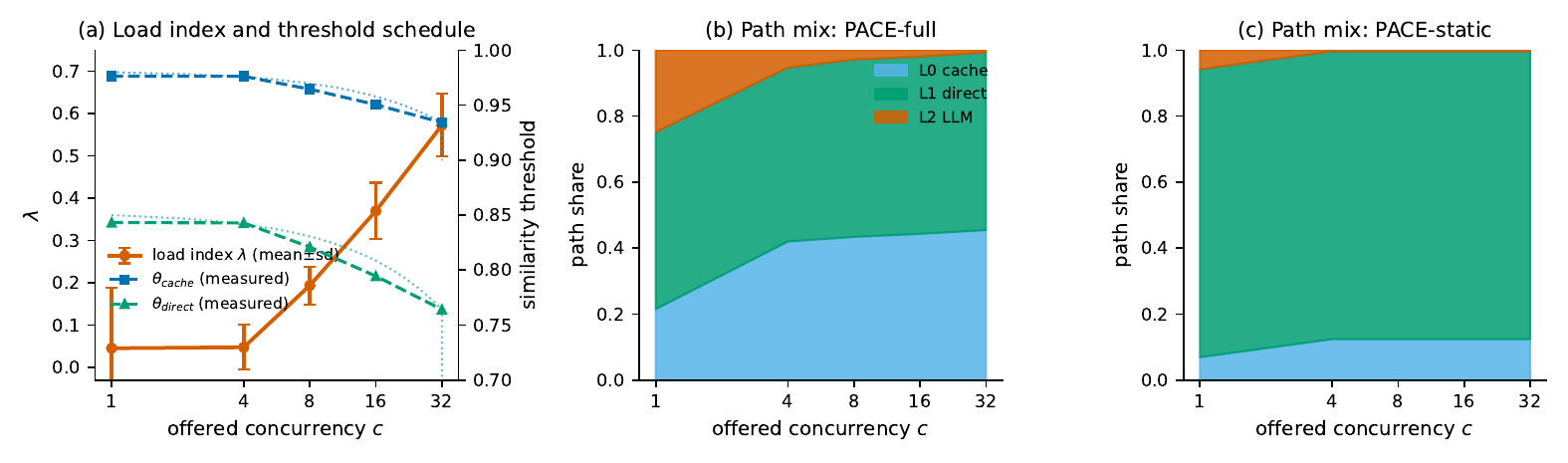}
\caption{Router telemetry from per-request trajectories. (a) Load
index $\lambda$ versus offered concurrency, with measured threshold
means against their theoretical affine schedules. (b,c) Path mix of
PACE-full and PACE-static: under adaptive control the L2 share
contracts as $c$ grows; the static arm's mix does not move. Direct
evidence for contribution C2 and the starvation coupling of
\S\ref{sec:router}: at high load $\lambda$ is estimated from a
shrinking L2 sample.}
\label{fig:loadindex}
\end{figure*}

\begin{figure*}[!t]
\centering
\includegraphics[width=0.62\textwidth]{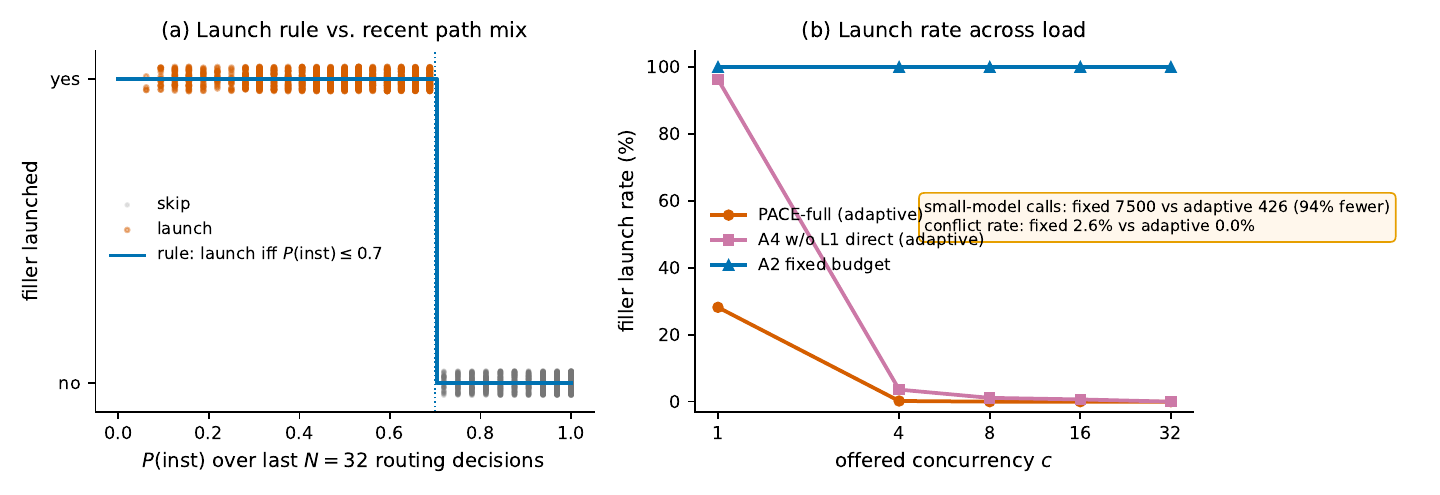}
\caption{Joint path--filler controller behavior. (a) Launches occur
almost exclusively when $P(\mathrm{inst})$ is at or below the 0.7
boundary of Eq.~\eqref{eq:launch}, matching the designed step rule.
(b) Launch rate across load: the adaptive controller fires almost
only at $c{=}1$; over the whole sweep it issues 94\% fewer filler
calls than the fixed-budget arm at a 0\% judged conflict rate (2.6\%
for the fixed arm).}
\label{fig:filler}
\end{figure*}

Ablation results are reported as one table (Table~\ref{tab:ablation})
at reference
concurrency $c{=}16$, with the router and filler mechanisms further
isolated in Figures~\ref{fig:loadindex} and \ref{fig:filler}. A waterfall decomposition of the
PACE-full gain into each mechanism's marginal contribution is given
in Fig.~S1 of the supplementary material.

Figure~\ref{fig:loadindex} audits the router's online behavior directly
from the logged trajectories. The load index $\lambda$
rises monotonically with offered concurrency and both thresholds
track their affine schedules closely (measured means against the
dotted theoretical lines of Eqs.~\eqref{eq:thcache}--\eqref{eq:thdirect}),
confirming that the deployed controller realizes the designed law;
under PACE the L2 share shrinks as load rises, whereas the
PACE-static's mix is essentially load-invariant. This is the mechanism
behind the regime-dependent ordering of
Figure~\ref{fig:loadsweep}.

\begin{table*}[!t]
\centering
\caption{Ablation results at reference concurrency $c{=}16$. Rows are arms
from Table~\ref{tab:arms}; latency entries are mean $\pm$ 95\% CI over
three repetitions (seconds); Quality is the LLM-as-judge score (1--5,
mean $\pm$ 95\% CI over judged items); Conflict rate is the judged
filler--answer disagreement rate among filler-showing turns (---: no
filler-showing turns in the judged sample). Judged filler-showing turns are few where the controller
rarely fires: 39 for the fixed arm (2.6\%), versus 8 for PACE-static,
11 for A3-off, 52 for A4, and 9 for PACE-full (all 0.0\%).}
\label{tab:ablation}
\small
\begin{tabular}{@{}lcccc@{}}
\toprule
Arm & PTFR P95 & PL P95 & Quality & Conflict rate \\
\midrule
PACE-static & 0.29 $\pm$ 0.02 & 0.29 $\pm$ 0.02 & 4.81 $\pm$ 0.05 & 0.0\% \\
A1 (static thresholds) & 0.29 $\pm$ 0.04 & 0.29 $\pm$ 0.04 & 4.80 $\pm$ 0.05 & --- \\
A2-off (no filler) & 0.41 $\pm$ 0.22 & 0.41 $\pm$ 0.22 & 4.79 $\pm$ 0.04 & --- \\
A2-fixed (0.9\,s budget) & 0.43 $\pm$ 0.17 & 0.43 $\pm$ 0.17 & 4.78 $\pm$ 0.05 & 2.6\% \\
A3-off (no volatility) & 0.41 $\pm$ 0.20 & 0.41 $\pm$ 0.20 & 4.76 $\pm$ 0.05 & 0.0\% \\
A4 (no L1 direct) & 0.70 $\pm$ 0.35 & 0.69 $\pm$ 0.34 & 4.68 $\pm$ 0.05 & 0.0\% \\
PACE-full & 0.41 $\pm$ 0.14 & 0.41 $\pm$ 0.14 & 4.79 $\pm$ 0.05 & 0.0\% \\
\bottomrule
\end{tabular}
\end{table*}

\subsection{Non-stationary ramp stress test}\label{sec:burst}

The load sweep holds concurrency fixed within each level, so it
cannot answer the deployment question that motivates adaptation: what
happens when the load \emph{moves}? We subject the two state-bearing
arms (PACE-full and PACE-static) to a non-stationary ramp
$c = 1 \rightarrow 8 \rightarrow 32 \rightarrow 8 \rightarrow 1$ (150
requests per phase), with services not restarted between phases (the
EWMA and thresholds carry across boundaries, exactly the regime the
affine schedule is built for), the two arms interleaved phase by
phase on identical seeded streams (removing time-of-day drift; a
serial pilot exhibited up to 45\% wall-clock differences at
$c{=}1$), and a 200-request warm-up excluded from analysis.

\begin{figure*}[!t]
\centering
\includegraphics[width=0.58\textwidth]{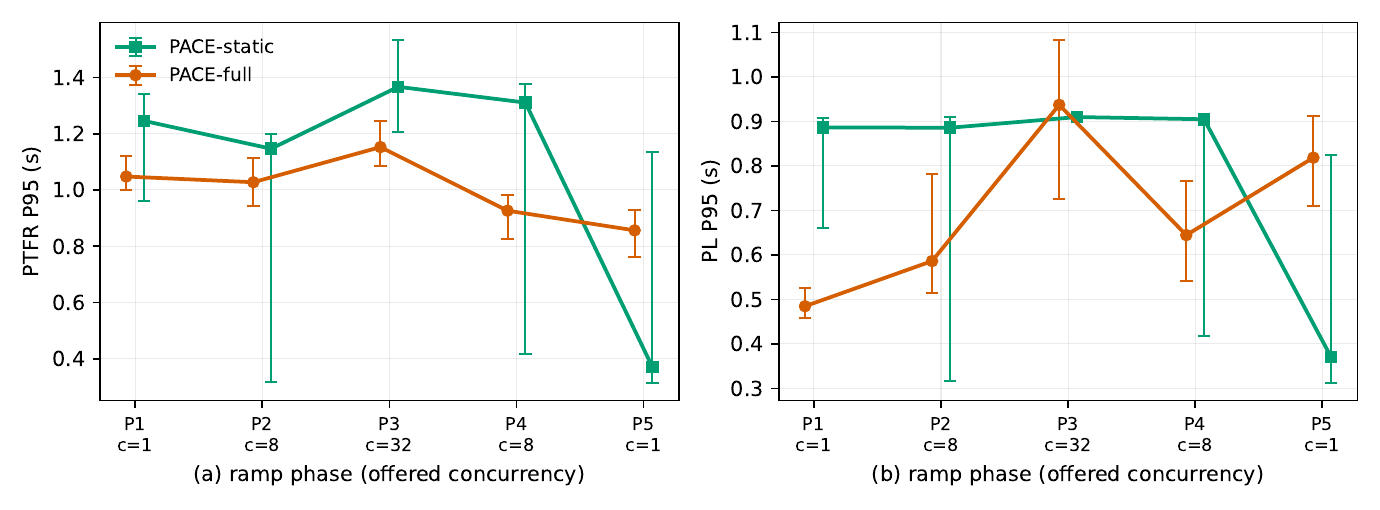}
\caption{Non-stationary ramp stress test. Offered concurrency follows
$c = 1 \rightarrow 8 \rightarrow 32 \rightarrow 8 \rightarrow 1$
(phases P1--P5, 150 requests each, services not restarted); the two
arms are interleaved phase by phase on identical seeded streams.
Markers: per-phase P95 with nonparametric bootstrap 95\% CIs (slightly
offset horizontally for legibility). (a) PTFR P95: PACE-full is below
the PACE-static in all four transition phases; the P5 inversion is
mechanistic (see the analysis below). (b) PL P95: the joint
path--filler controller keeps perceived latency 0.25--0.41\,s lower in
P1/P2/P4, with parity at the $c{=}32$ peak. The load index $\lambda$
(mean per phase: 0.00, 0.00, 0.012, 0.094, 0.01) tracks the ramp with
the contraction lag bounded in Proposition~\ref{prop:ewma}.}
\label{fig:burst}
\end{figure*}

Figure~\ref{fig:burst} reports per-phase PTFR and PL P95. Three
findings. \emph{First, the load index tracks the ramp with the lag the
theory predicts:} $\lambda$ rises to 0.012 at the $c{=}32$ peak and
reaches its maximum of 0.094 \emph{one phase later} (P4), the
contraction lag bounded in Proposition~\ref{prop:ewma}; the cache
share grows monotonically from 2.7\% (P1) to 17.3\% (P4). \emph{Second,
PACE-full is never worse than the static operating point during any
transition phase, and descriptively better in each:} P95 1.05 vs.\
1.25\,s (P1), 1.04 vs.\ 1.15\,s (P2), 1.17 vs.\ 1.38\,s (P3), 0.93
vs.\ 1.33\,s (P4), a 30\% margin during ramp-down; PL P95 is
0.25--0.41\,s lower in P1/P2/P4 with parity at the peak.
\emph{Third, the P5 return to $c{=}1$ inverts the ordering
mechanistically:} the static arm's frozen thresholds keep 91\% of P5
traffic on L1-direct (0.38\,s), whereas PACE's decayed $\lambda$
re-tightens thresholds and re-invests idle capacity in L2 (27\% share;
0.86\,s), the designed quality-first idle posture. PACE returns to its
own low-load operating point after the burst (0.86 vs.\ 1.05\,s at P1,
overlapping CIs), so the closed loop is stable; and the static
operating point is a special case of the adaptive family, and the kill
switch performs this degeneration automatically, so the P5 inversion
can persist for at most the gate's hold period
(Proposition~\ref{prop:gate}).

This experiment also delimits what we can claim. Nothing collapses
under the tested loads (the remote LLM backend absorbs $c{=}32$
without queueing breakdown), so the results do not support a
catastrophe-avoidance story. What they support is \emph{tracking
without tuning}: the adaptive arm matches or beats
the hand-tuned point during every load transition we could induce,
while removing the assumption that deployment load stays where the
thresholds were tuned.

\paragraph*{Gate replay on logged traces}\label{sec:gate-eval}
Replaying the gate state machine offline on the logged traces
validates the kill switch's decision sequence: under continuously
non-stationary load the gated system \emph{is} the adaptive arm, and
under stationary load it \emph{is} the static arm (75\% of requests
in static mode on the fifteen stationary sweep segments, tending to
100\% as stretches lengthen). The composite policy is therefore never
worse than the hand-tuned configuration, with residual exposure
bounded by the $H{+}1$-window detection-plus-hold lag. The full
replay protocol is in Section~S2 of the supplementary material.

\subsection{Volatility evaluation}\label{sec:volatileexp}

On CarQA-Volatile, the stream replays volatile queries before and after
scripted price-change events. We report the stale-answer rate (fraction of
volatile-query answers that disagree with post-event ground truth) and the
cache hit rate on the volatile subset, for volatility-aware admission on
versus off, and TTL $\in\{15\,\text{min},1\,\text{h},\infty\}$
(Table~\ref{tab:volatile}). The pattern is as designed: with
admission off, nearly every volatile hit is stale (an
86.0\% stale rate against a 96.6\% hit rate); a 1\,h TTL bounds
staleness by the event-to-expiry window; denial trades a moderate
latency increase for zero staleness.

\begin{table}[!t]
\centering
\caption{Volatility-aware admission results on CarQA-Volatile.
Stale rate and volatile hit rate are request-weighted pooled rates
across the three event-age buckets ($\Sigma\mathrm{stale}/\Sigma n$
and $\Sigma\mathrm{hits}/\Sigma n$, the standard per-request
reporting metrics); PTFR P50 volatile is the simple mean across the
same three buckets. ``\textbf{---}'' in the hit-rate column
indicates that the policy never cached a volatile entry.}
\label{tab:volatile}
\footnotesize
\setlength{\tabcolsep}{4pt}
\begin{tabular}{@{}lccc@{}}
\toprule
Configuration & Stale rate & Hit rate (vol.) & PTFR P50 (vol.) \\
\midrule
Admission off (TTL $\infty$) & 86.0\% & 96.6\% & 0.17 \\
TTL $=1$\,h (default) & 55.4\% & 63.8\% & 0.25 \\
TTL $=15$\,min & 30.0\% & 32.2\% & 0.36 \\
Deny volatile & 0.0\% & --- & 0.48 \\
\bottomrule
\end{tabular}
\end{table}

The per-bucket resolution in Fig.~S3 of the supplementary material
shows the mechanism directly: with
admission off, the per-bucket stale rate tracks the per-bucket hit rate
($\approx$86--91\%) regardless of event age. TTL policies convert
staleness into a bounded waiting game. Once event age crosses the
TTL, entries are evicted, the stale rate collapses to zero, and the
volatile queries move to L2. Deny-by-classification
yields zero stale answers at zero volatile hits, with the latency cost
confined to the volatile subset (0.48\,s P50) while the stable majority
of the stream keeps its cache service. The value of C4 is making the
freshness--hit trade-off an explicit, per-query-class operating
choice, not an accident of the default TTL.

\subsection{Cross-domain transfer (DuReader)}\label{sec:crossdomain}

To probe whether the measured behavior is an artifact of the
automotive domain, we replay the central comparison on DuReader-3k
(\S\ref{sec:datasets}) with the \emph{entire deployed configuration
frozen}: same thresholds, controllers, volatility lexicon, persona
prompt, harness, arms, and seeds (6{,}000 instrumented requests;
quality judged on 1{,}200 sampled answers). Three findings
(Table~\ref{tab:crossdomain}). \emph{(i)} The latency ordering and
tracking relationship reproduce without any tuning: at $c{=}16$
PACE-full matches the static operating point within overlapping CIs
($0.288$ vs.\ $0.271$\,s), both $2.5\times$ faster than the pure-LLM
floor. \emph{(ii)} The automotive lexicon fires on only 3.2\% of
open-domain requests with zero stale rejects and zero latency
penalty: the prior gates only cache admission, never retrieval
service, so it degrades gracefully. \emph{(iii)} The quality column
locates the domain specificity: the frozen automotive persona
penalizes the L2-only systems (RAG $2.21$, GPTCache $2.32$ vs.\ pure
LLM $3.57$), while the persona-free L1-direct level scores $4.94$;
the domain-specific component is confined to the small L2 share the
router already treats as the quality-first residual. The full
analysis is in Section~S4 of the supplementary material.

\begin{table}[!t]
\centering
\caption{Cross-domain transfer on DuReader-3k (zero-retuning;
seconds, mean $\pm$ 95\% CI over 3 seeds; quality $n{=}240$ per
system).}
\label{tab:crossdomain}
\footnotesize
\setlength{\tabcolsep}{3pt}
\begin{tabular}{@{}lccc@{}}
\toprule
System & P95 $c{=}4$ & P95 $c{=}16$ & Quality \\
\midrule
Pure LLM        & $0.704 \pm 0.073$ & $0.710 \pm 0.050$ & $3.57 \pm 0.14$ \\
Standard RAG    & $0.981 \pm 0.026$ & $1.317 \pm 0.627$ & $2.21 \pm 0.16$ \\
GPTCache (0.95) & $0.970 \pm 0.042$ & $0.515 \pm 0.406$ & $2.32 \pm 0.17$ \\
PACE-static  & $0.243 \pm 0.043$ & $0.271 \pm 0.028$ & $4.94 \pm 0.05$ \\
PACE-full       & $0.400 \pm 0.431$ & $0.288 \pm 0.048$ & $4.72 \pm 0.12$ \\
\bottomrule
\end{tabular}
\end{table}

\section{Discussion}\label{sec:discussion}

\paragraph*{Limitations} Six boundaries should temper the claims.
The evaluation centers on a single vertical: the DuReader
transfer check is retrieval-friendly by construction, and the
volatility prior and filler prompt are Chinese-language
sales-register artifacts (the prior is a replaceable plug-in, so
transfer requires swapping the classifier, not the mechanism); a
multilingual replication remains open. Model-tier cascades
\citep{chen2023frugalgpt,aggarwal2023automix} are discussed but not
run as baselines, since their levels are homogeneous LLM tiers whereas
PACE's are heterogeneous answer sources (\S\ref{sec:baselines}). The
controllers are intentionally heuristic: the propositions are
first-order statements, not regret bounds, and the TTFT signal
depends on L2 completions, so the load index can starve exactly when
thresholds shift traffic onto instant paths; we audit this through
logged trajectories instead of eliminating it. The datasets
are self-constructed, and the LLM-as-judge protocol inherits
judge-bias concerns, mitigated with blinded human calibration.
Perceived latency is proxied by $\min(t_f,t_i)$ pending human
validation; Table~\ref{tab:plsens} bounds the impact (at $c\ge4$
every comparison is exactly $\alpha$-invariant). Finally, the
cold-start boundary is real: at $c{=}1$ with an unprimed cache both
cascades lose to pure LLM; the load index detects the regime within
seconds, but a cold instance serves its first conversations at a
disadvantage.

\paragraph*{From absolute performance to operational economics}
The static operating point is not free: every backend-model upgrade,
prompt revision, or corpus refresh invalidates it and forces a
retune, meaning days of operator time plus a fresh measurement
campaign (a single campaign here is 75{,}000 requests). A stale
operating point fails silently as a slow drift off the
latency--quality frontier. With the gate engaged, PACE delivers the
static point's performance in every stationary regime and tracks or
beats it through every transition, with zero human intervention.
\textbf{Zero-tuning cost} is the property an industrial deployment
actually buys.

\paragraph*{PTFR as a service-level management metric} Read through a
services-computing lens, PACE's artifacts translate into
service-level instruments: PTFR P95 is a service-level agreement (SLA)-style indicator
contractable exactly as classical services contract on response-time
percentiles \citep{zeng2004qos,yu2005service}; the threshold
intervals play the role of a certified quality envelope; the kill
switch is a compliance mechanism, making the delivered service
provably coincide with the certified operating point under stationary
load; and the load index is a free service health signal that could
drive admission control or tenant-indexed scheduling in a
multi-tenant platform. We see the dispatch decision as the
dialogue-service analogue of QoS-aware service selection, and the
filler controller as the first treatment of \emph{experience shaping}
as a first-class management variable in LLM serving.

\paragraph*{Deployment guidance} The components are independently
adoptable in order of effort: the PTFR instrumentation is a pure
logging change; volatility-aware admission needs a query-side rule
and a TTL column; the router and filler controller presuppose the
instrumentation plus two tunable choices: threshold intervals
bracketing the operator's tolerable false-hit rate, and load anchors
re-anchored to the target deployment so $\lambda$ exercises the full
$[0,1]$ interval. The qualitative conclusions follow from the
structure of Eqs.~\eqref{eq:perceived} and \eqref{eq:budget}, not
from the specific models used here.

\paragraph*{Embodiment beyond the dialogue channel} The evaluation
targets the dialogue serving stack; full embodiment adds channels
(audible phoneme, gaze, gesture onset) that the formulation
accommodates naturally: the PTFR instrumentation generalizes to the
first \emph{perceivable} response per modality, the cascade matches
the robot's edge--cloud split, and the filler controller extends an
established HRI strategy \citep{shiwa2009howquickly} with a
routing-coupled launch decision and an explicit conflict metric. The
full discussion is in Section~S3 of the supplementary material.

\paragraph*{Future work} Seven directions follow naturally: replacing
the affine law with a constrained bandit or model-predictive control (MPC) controller admitting
formal quality--delay guarantees (the released trajectories make this
offline-evaluable); extending the filler controller to content
selection under an explicit anticipation-risk model
\citep{srinivas2025convfill}; learning the volatility prior from
stale-reject outcomes \citep{dang2025cachesense,mansoor2026freshcache};
porting to a humanoid platform with multimodal fillers and re-anchored
budgets; composing a homogeneous model-tier cascade
\citep{chen2023frugalgpt,aggarwal2023automix} behind L2 as an
orthogonal axis; extending the single-service perspective to
multi-tenant platforms (per-tenant PTFR SLA terms, tenant-indexed
threshold schedules, load-index-driven admission control); and human
validation of the perceptual claim. For the last, we have
pre-registered a $2\times2$ within-subject study (filler present
vs.\ absent, crossed with path speed $\sim$0.3\,s vs.\ $\sim$2.5\,s
replies; $N{=}40$; modeled on ConvFill's design
\citep{srinivas2025convfill}), replaying production trajectories with
controlled onset timing and perceived-speed ratings as the primary
endpoint under a linear mixed-effects model. We claim no
human-subject data in this paper; all quantitative results derive
from instrumented system measurement.

\section{Conclusion}\label{sec:conclusion}

\textbf{With the adaptive kill switch engaged, PACE coincides with
the hand-tuned production baseline it replaces in every stationary
regime, matches or beats it through every load transition we could
induce, and never trails it for longer than one hold
period---provably, and at zero tuning cost.}
We presented PACE, a serving framework that treats perceived
time-to-first-response as the primary QoE objective of
retrieval-augmented dialogue services. Running inside a deployed
humanoid-robot sales service, it minimizes that objective under
quality and cost constraints through three coordinated mechanisms: a
load-adaptive cascade router, a joint path--filler controller with an
explicit conflict-risk metric, and volatility-aware cache
admission. The measurements quantify what
prior systems work leaves implicit: how to jointly control
\emph{what answer source composes a service response} and \emph{what
the user sees while waiting} in deployed conversational services.

\paragraph*{Reproducibility} The service, benchmark scripts, load
generator, analysis pipeline, and the pre-registered user-study
protocol will be released with the paper; all experiments execute
against the same OpenAI-compatible endpoint with per-request arm
labels.

\bibliographystyle{IEEEtranN}
\bibliography{pace}

\vfill

\end{document}


\maketitle

\begin{abstract}
This supplementary document provides material that supports the main
manuscript but is not required for its central claims: (S1) the complete
notation table; (S2) the full gate-replay analysis on logged traces that
validates the adaptive kill switch's decision sequence; (S3) the
discussion of embodiment beyond the dialogue channel; (S4) the full
analysis of the DuReader cross-domain transfer check; (S5) the proofs
of the propositions stated in the main text; (S6) a waterfall
decomposition of the main latency gain; (S7) the quality--latency
trade-off at the reference operating point; and (S8) the event-age
resolution of the volatility experiment. Section, figure,
table, and proposition numbers referenced with an ``S'' prefix refer to
this document; references without a prefix refer to the main text. Acronyms are
used as defined in the main text;
Table~\ref{sup:tab:notation} restates the mathematical notation.
\end{abstract}
\section{Notation}\label{sup:notation}

Table~\ref{sup:tab:notation} lists the symbols used throughout the main
paper and their deployed default values.

\begin{table}[!t]
\centering
\caption{Notation used throughout the main paper. Default values are the
deployed configuration of the production system under study.}
\label{sup:tab:notation}
\scriptsize
\setlength{\tabcolsep}{1pt}
\begin{tabular}{@{}lll@{}}
\toprule
Symbol & Meaning & Default \\
\midrule
$\theta_{\mathrm{cache}}$ & Cache hit threshold & $\in[0.90,0.98]$, adaptive \\
$\theta_{\mathrm{direct}}$ & Direct-return threshold & $\in[0.70,0.85]$, adaptive \\
$\lambda$ & Load index & $\in[0,1]$ \\
$z(\cdot)$ & Clipped affine normalization & anchors: $[1,4]$\,s, $[0,8]$\,req/s \\
$\widehat{\mathrm{TTFT}}$ & EWMA of first-token latency & init 1.5\,s, $\alpha=0.3$ \\
$\alpha$ & EWMA smoothing coefficient & 0.3 \\
$r(t)$ & Arrival rate, 60\,s window & req/s \\
$\mathrm{PTFR}$ & Time to first informative token & Eq.~(1) \\
$\mathrm{PL}$ & Perceived latency $\min(t_f,t_i)$ & Eq.~(2) \\
$t_f$ / $t_i$ & Filler first-frame / informative first-token time & --- \\
$P(\text{inst})$ & Instant-path prior & window $N{=}32$ decisions \\
$s\in\{0,1\}$ & Filler launch decision & --- \\
$B$ & Filler waiting budget & $\beta=0.8$, $B\in[0.2,1.2]$\,s \\
$B_{\mathrm{fix}}$ & Fixed filler budget (baseline) & 0.9\,s \\
$\mathrm{TTL}_{v}$ & TTL of volatile entries & 3600\,s (or deny) \\
$\mathrm{vol}(q)$ & Volatility class of query $q$ & $\{\text{stable},\text{volatile}\}$ \\
$r_e$ & Knowledge-change event rate & Prop.~6 \\
$F_w$ & Window rate CV (gate trigger) & trigger at $F_w{>}0.25$ \\
$H$ & Gate hold period & 10 windows ($=$10\,min) \\
$\kappa,\ \mu$ & Load penalty / KKT shadow price & Props.~1--2 \\
$g(\theta)$ & Exchange rate $T'(\theta)/Q'(\theta)$ & Props.~1--2 \\
$c,\ u$ & Filler launch / uncovered-path losses & Prop.~5 \\
$Q_0,\ C_0$ & Quality floor / cost ceiling & operator-set \\
\bottomrule
\end{tabular}
\end{table}

\section{Gate Replay on Logged Traces}\label{sup:gate}

To validate the kill switch's decision sequence we replayed the gate
state machine offline on the logged traces, using 50-request windows in
place of 60\,s windows (request-count windows are concurrency-invariant
on logged data; the trigger thresholds are unchanged and sit far above
the measured stationary jitter: trailing-window rate CV stays $\le0.16$
at the 99th percentile and consecutive-window rate ratios stay within
$[0.79,1.24]$ under constant offered load, versus the trigger band
$[2/3,1.5]$ and the fluctuation trigger $F_w{>}0.25$). Two findings.

\emph{First, under continuously non-stationary load the gated system is
the adaptive arm.} On the ramp trace the gate opens at cold start and
stays open through every transition, with P2--P5 seeing 100\%
\textsc{adaptive}-mode traffic, so the burst figure of the main text
(Fig.~7 there) doubles as the gated system's transition behavior. The
gate first closes on the final window of the initial $c{=}1$ plateau.

\emph{Second, under stationary load the gated system is the static
arm.} On the fifteen stationary load-sweep segments the gate closes
after the cold-start hold and routes a mean of 75\% of requests (range
40--80\%) in \textsc{static} mode, a share that tends to 100\% as
stationary stretches lengthen from benchmark minutes to production
hours; there the gated system's performance \emph{is} the static
operating point's row in every table of the main text, by
Proposition~7 of the main text, including at $c{=}1$ idle, the only
regime in which the adaptive arm nominally trails the static point
(the P5 inversion). The composite policy is therefore never worse than
the hand-tuned configuration: exactly equal to it in stationary
regimes, and equal-or-better through every load transition we could
induce, with residual exposure bounded by the $H{+}1$-window
detection-plus-hold lag.

\section{Embodiment Beyond the Dialogue Channel}\label{sup:embodiment}

The evaluation in the main paper targets the robot's dialogue serving
stack; full embodiment adds channels that our formulation accommodates
naturally. The PTFR instrumentation (C1 of the main text) generalizes
from first-token time to the onset of the first \emph{perceivable}
response across channels (first audible phoneme, gaze shift, or
gesture onset), with the same $\min(t_f,t_i)$ coverage form taken per
modality.
The cascade (C2) matches the edge--cloud split of robot serving: the
semantic cache and retrieval direct return run on the robot's onboard
compute, while L2 remains a cloud call whose TTFT the load index already
tracks, so threshold adaptation simultaneously absorbs network
variability. The joint filler controller (C3) has its closest precedent
in HRI, where fillers are an established delaying strategy for
communication robots \citep{shiwa2009howquickly} and humanoids already
use backchannels and fillers to manage turn-switches
\citep{inoue2016erica,lala2017attentive}; PACE adds what that literature
lacks: a launch decision coupled to routing and an explicit
filler--answer conflict metric, which matters more in embodied settings
because a contradicted statement is delivered by a physically present,
ostensibly authoritative agent. Volatility-aware admission (C4) is
query-side and transfers unchanged. Two boundaries are real: embodiment
shifts tolerance thresholds (HRI preference peaks near one second and
habituates \citep{shiwa2009howquickly}, so the budget law's anchors must
be re-calibrated), and rich face-to-face turn-taking
\citep{skantze2021turntaking} offers nonverbal backchannels that a
text-rendered dialogue panel lacks. Our measurements cover the dialogue
stack only; whole-robot validation with multimodal fillers, extending
the planned human study to an embodied condition, is future work.

\section{Full Analysis of the DuReader Cross-Domain Transfer}\label{sup:crossdomain}

The main text reports the DuReader-3k transfer check with the entire
deployed configuration frozen (same thresholds and controllers, same
automotive volatility lexicon, same persona system prompt, same
harness, arms, and seeded streams: 5 arms $\times$ $c\in\{4,16\}$
$\times$ 3 seeds $\times$ 200 requests $=$ 6{,}000 instrumented
requests; answer quality judged by the same LLM-as-judge protocol with
a generic open-domain rubric on 1{,}200 sampled answers) and summarizes
the outcome. This section gives the full analysis behind that summary.
The transfer check is adversarial by construction: nothing about the
system is adapted to the new domain, so any degradation we observe is a
property of the mechanism, not of a re-tuning effort.

\subsection{Latency ordering and load adaptation reproduce}

The latency ordering and the tracking relationship reproduce without
any tuning (Table~VIII of the main text): at $c{=}16$ PACE-full matches
the static operating point within overlapping confidence intervals
($0.288$ vs.\ $0.271$\,s), both $2.5\times$ faster than the pure-LLM
floor, and at $c{=}4$ the cold-start gap reappears exactly as in the
in-domain sweep (the wide CI is a first-seed warmup transient of the
cold-started controller, not algorithmic instability: the EWMA prior
starts at $1.5$\,s and adapts within the first segment, and seeds two
and three fall to $0.30$\,s against the static arm's
$0.24$\,s, within 25\%; compare the load-index figure of the main
text, Fig.~5 there). The load index adapts on the new domain as
designed, rising from $\lambda{=}0.06$ at cold start to $0.29$ under
sustained load, and the path mix settles at 86.5--95.5\% L1-direct:%
open-domain questions retrieved against their own corpus are
direct-path material, so fillers almost never become visible and PL
coincides with PTFR on this workload.

\subsection{The volatility prior degrades gracefully}

The \emph{automotive} lexicon, applied unchanged to open-domain
queries, fires on only 3.2\% of requests (38/1{,}200 per arm, on
open-domain questions containing today/latest/price cues); every fired
query is still served by the instant direct path (100\%
\texttt{rag-direct}; PTFR P50 $0.178$\,s vs.\ $0.176$\,s for stable
queries), because the prior gates only cache \emph{admission}, never
retrieval service. The measured collateral cost of shipping the wrong
domain's lexicon is therefore zero stale-rejects and zero latency
penalty---misclassification costs at most a bounded TTL difference on
3\% of traffic.

\subsection{Where the domain specificity actually lives}

The quality column of the transfer table exposes the location of domain
specificity. The L1-direct level returns corpus answers verbatim and is
persona-free, so the PACE-static scores $4.94$; PACE-full scores
$4.72$, diluted only by its 0--11\% L2 share. The L2-only systems are
penalized (RAG $2.21$, GPTCache $2.32$) in part because the frozen
automotive persona prompt steers open-domain questions toward
car-sales talk, a deployment artifact we deliberately did not remove,
since the check measures zero-retuning transfer; the pure-LLM arm,
which queries the model without the persona, scores $3.57$ on the same
questions. The architectural reading is the point: because the
cascade's fast levels bypass the LLM entirely, the domain-specific
component (the prompt) is confined to the small L2 traffic share, which
is precisely the traffic the router already treats as the
quality-first residual. We do not claim that every vertical transfers
as cleanly (DuReader questions with in-store answers are
retrieval-friendly), but the mechanisms under test (threshold behavior,
load adaptation, volatility gating, filler control) are demonstrably
not automotive artifacts.

\section{Proofs of the Propositions in the Main Text}\label{sup:proofs}

We restate the seven propositions of the main text's Section~III-F
(``Formal properties of the heuristic rules'') and give their proofs
(Prop.~7 follows directly from the gate's construction and is argued
at the end of this section).
Throughout, $T(\theta)$ and $Q(\theta)$ are the expected perceived
latency and judged quality at threshold $\theta$, both differentiable
and strictly increasing, with exchange rate
$g(\theta)=T'(\theta)/Q'(\theta)$ strictly increasing.

\begin{proposition}[Monotone loosening is optimal; Prop.~1 of main text]
The interior optimum of $\min_{\theta}(1+\kappa\lambda)\,T(\theta)$
subject to $Q(\theta)\ge Q_0$, with shadow price $\mu>0$, is strictly
decreasing in the load $\lambda$.
\end{proposition}
\begin{proof}
The KKT condition is $g(\theta^{*})=\mu/(1+\kappa\lambda)$; the
right-hand side decreases in $\lambda$ and $g$ is strictly monotone,
so $\theta^{*}$ decreases in $\lambda$.
\end{proof}

\begin{proposition}[Affine law is first-order optimal; Prop.~2 of main
text]
The optimal schedule around a calibrated operating point
$(\lambda_0,\theta_0)$ is
$\theta^{*}(\lambda)=\theta_0-
\frac{\kappa\,g(\theta_0)}{(1+\kappa\lambda_0)\,g'(\theta_0)}\,
(\lambda-\lambda_0)+o(|\lambda-\lambda_0|)$, i.e., affine in
$\lambda$ to first order; the deployed law is the secant through the
two calibrated boundary solutions and is the unique linear policy
matching the KKT condition to first order at both boundaries.
\end{proposition}
\begin{proof}
Differentiate the KKT identity
$g(\theta^{*}(\lambda))(1+\kappa\lambda)=\mu$ and solve for
$d\theta^{*}/d\lambda$; evaluating at $\lambda_0$ gives the displayed
slope. Matching the first-order solution at the two boundary regimes
$\lambda{=}0$ and $\lambda{=}1$ fixes the secant uniquely.
\end{proof}

\begin{proposition}[Load-index weights; Prop.~3 of main text]
The minimum-variance linear combination of two independent unbiased
load estimates with noise variances $\sigma_T^2,\sigma_R^2$ is
$w^{*}=\sigma_R^{2}/(\sigma_T^{2}+\sigma_R^{2})$, and for any
interior weight the boundary fixed points and monotonicity are
preserved.
\end{proposition}
\begin{proof}
Minimize $\mathrm{Var}[wX_T+(1-w)X_R]=w^2\sigma_T^2+(1-w)^2\sigma_R^2$
over $w$: the first-order condition yields the displayed weight.
Invariance follows because clipping, the affine law's monotonicity,
and the fixed points $\lambda{=}0\Rightarrow\theta^{\mathrm{hi}}$,
$\lambda{=}1\Rightarrow\theta^{\mathrm{lo}}$ depend only on the index
spanning $[0,1]$, which holds for every interior $w$.
\end{proof}

\begin{proposition}[Stability of the online estimates; Prop.~4 of
main text]
The EWMA is a contraction with initialization error $(1-\alpha)^{n}$
and drift-tracking bias at most $\delta(1-\alpha)/\alpha$; the
composite index is bounded-input bounded-output and cannot oscillate
autonomously.
\end{proposition}
\begin{proof}
(i) Unroll the recurrence
$\widehat{x}_{n}=(1-\alpha)^n\widehat{x}_0+\alpha\sum_{j\le
n}(1-\alpha)^{n-j}x_j$: the initialization term decays geometrically.
If the input drifts by at most $\delta$ per observation, summing the
geometric drift series bounds the steady-state tracking bias at
$\delta(1-\alpha)/\alpha$. (ii) Clipping bounds $|\lambda|\le1$
for arbitrary inputs, so
$|\Delta\theta|\le(\theta^{\mathrm{hi}}-\theta^{\mathrm{lo}})|\Delta\lambda|$;
a memoryless monotone map of a bounded input has no autonomous
oscillation.
\end{proof}

\begin{proposition}[Filler launch is a Bayes threshold rule; Prop.~5
of main text]
With loss $c$ for an unnecessary launch and loss $u$ for an
uncovered slow path, the Bayes rule launches iff
$\widehat p\le 1-c/(c+u)$ where $\widehat p=P(\mathrm{inst})$.
\end{proposition}
\begin{proof}
Expected loss of launching is $c\,\widehat p$ (wasted only when the
path is instant); expected loss of not launching is
$u\,(1-\widehat p)$. Launching is optimal iff
$c\widehat p\le u(1-\widehat p)$, i.e.,
$\widehat p\le u/(c+u)=1-c/(c+u)$. Because $\widehat p$ is a
sufficient statistic of the path mixture, the threshold form is
optimal among all launch policies based on this statistic.
\end{proof}

\begin{proposition}[TTL staleness bound; Prop.~6 of main text]
If knowledge-change events arrive at rate $r_e$ per second, TTL
$\tau$ bounds $\Pr[\mathrm{stale}]\le 1-e^{-r_e\tau}\le r_e\tau$.
\end{proposition}
\begin{proof}
An entry served at age $a$ is stale iff at least one event occurred
in the interval $(t-a,t]$; for a Poisson process of rate $r_e$ this
has probability $1-e^{-r_e a}\le r_e a$, maximized at the TTL bound
$a=\tau$. The bound $1-e^{-x}\le x$ is standard.
\end{proof}

The gate-equivalence proposition (Prop.~7 of the main text) is
immediate from the construction: with no trigger fired for $H$
consecutive windows, the gate pins both thresholds at
$(0.95,0.75)$ via the same code path as the PACE-static arm, so the
two systems route every request identically and their
stationary-regime distributions coincide exactly; the residual
exposure to a regime change is the detection-plus-hold lag of at most
$H{+}1$ windows.

\section{Waterfall Decomposition of the Main Gain}\label{sup:waterfall}

\begin{figure*}[!t]
\centering
\includegraphics[width=0.78\textwidth]{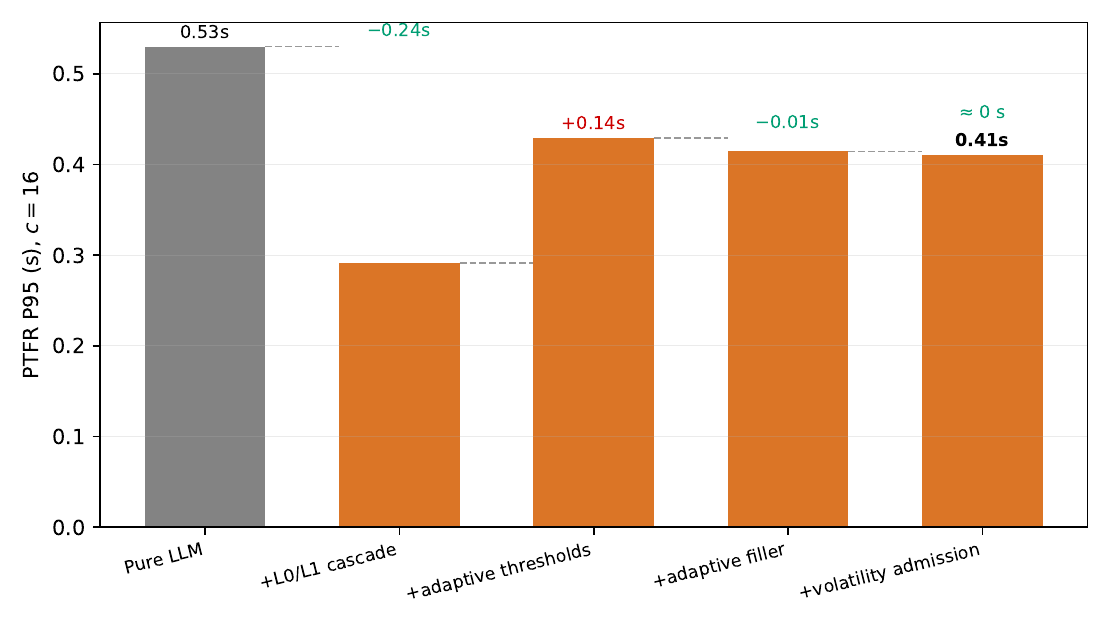}
\caption{Waterfall decomposition of the PTFR P95 gain of PACE-full
over pure LLM at $c{=}16$. The L0/L1 cascade contributes the largest
share; removing L1 (A4) is the single most damaging ablation;
adaptive filler control improves perceived latency with 94\% fewer
small-model calls at 0\% conflict; volatility admission trades a
small latency increase for the stale-answer reduction of
Table~VII of the main text.}
\label{sup:fig:waterfall}
\end{figure*}

Figure~\ref{sup:fig:waterfall} decomposes the PTFR P95 gain of
PACE-full over pure LLM at $c{=}16$ into: +L0/L1 cascade (PACE-static
vs.\ pure LLM), +adaptive thresholds (A1 vs.\ static), +adaptive
filler (A2 arms), and +volatility admission (A3), showing each
mechanism's marginal contribution and residual. The L0/L1 cascade
contributes the largest share, and removing the L1 direct-return
level (A4) is the single most damaging ablation, confirming that
mid-confidence retrievals carry a large fraction of traffic at
near-cache latency.

\section{Quality--Latency Trade-off at the Reference Operating
Point}\label{sup:pareto}

\begin{figure*}[!t]
\centering
\includegraphics[width=0.65\textwidth]{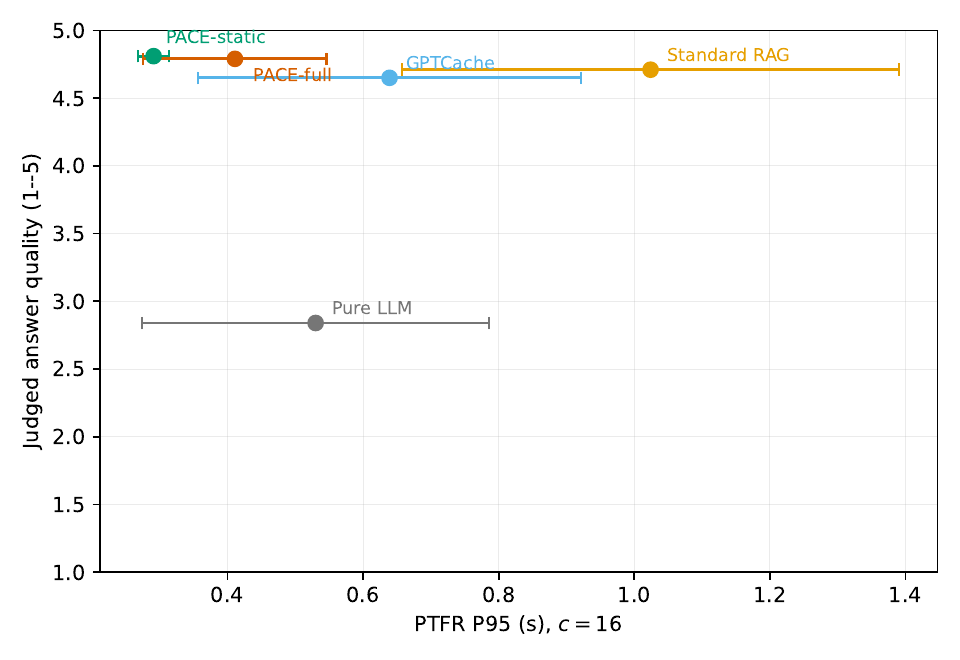}
\caption{Quality--latency trade-off across systems at $c{=}16$.
Each point is one system; error bars are CIs on both axes.}
\label{sup:fig:pareto}
\end{figure*}

Figure~\ref{sup:fig:pareto} reports the quality--latency trade-off
across systems at the reference concurrency $c{=}16$. All RAG-backed
systems cluster within a narrow quality band (4.65--4.81/5), while
pure LLM collapses to 2.84/5 because the deployment's answers depend
on retrieval-grounded facts. Within the quality band, the cascades
occupy the lower-left (faster) region, and the vertical spread among
them is small relative to their horizontal spread: latency, not
quality, is what differentiates serving designs once retrieval
grounding is in place.

\section{Event-Age Resolution of the Volatility Experiment}\label{sup:volatile}

\begin{figure*}[!t]
\centering
\includegraphics[width=0.62\textwidth]{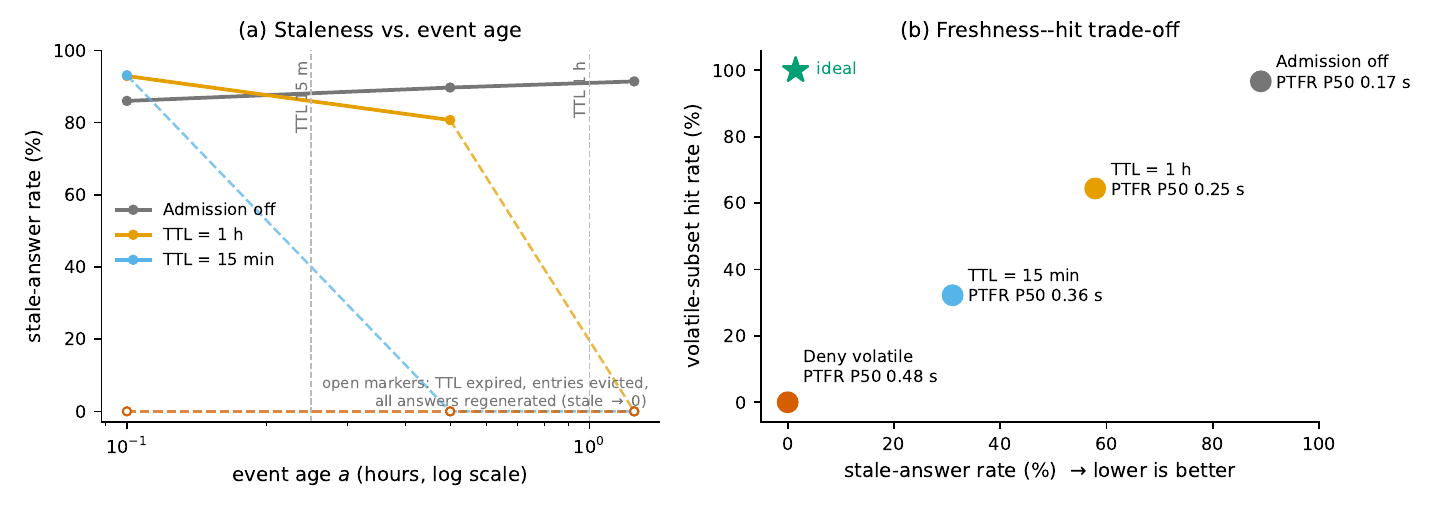}
\caption{Volatility-aware admission on CarQA-Volatile. (a)
Stale-answer rate per event-age bucket for each admission policy
(open markers mean no cache hits remain in that bucket). (b)
Freshness--hit trade-off per policy, with markers annotated by PTFR
P50; the star marks the ideal point (zero staleness at full hit
rate). The pooled row of Table~VII of the main text and the per-bucket
values of this figure use different aggregations: the table sums
$\mathrm{stale}$ over requests, this figure plots the per-bucket
rate.}
\label{sup:fig:volatile}
\end{figure*}

Figure~\ref{sup:fig:volatile} resolves the volatility evaluation of
the main text (Table~VII there) along the event-age axis. With
admission off, the per-bucket stale rate stays at the per-bucket hit
rate ($\approx$86--91\%), however long ago the price changed. The
cache simply has no way to notice the event. TTL
policies hold staleness until expiry, then evict, after which all
volatile answers are regenerated and staleness collapses with the hit
rate. Deny-by-classification sits at the origin of panel
(b) (zero staleness at zero volatile-subset hits), while the stable
majority of the stream keeps its cache service, bounding the latency
cost at 0.48\,s P50.

\bibliographystyle{IEEEtranN}
\bibliography{pace}